\documentclass{article}

     \usepackage[preprint, nonatbib]{neurips_2019}

\usepackage[utf8]{inputenc} 
\usepackage[T1]{fontenc}    
\usepackage{hyperref}       
\usepackage{url}            
\usepackage{booktabs}       
\usepackage{amsfonts}       
\usepackage{nicefrac}       
\usepackage{microtype}      
\usepackage{epsf, latexsym}
\usepackage{xspace, amsmath, amssymb, hyperref}
\usepackage{color}
\usepackage{graphicx}
\usepackage{epstopdf}
\usepackage{subcaption}
\usepackage{enumitem}
\usepackage{tikz}
\usepackage{bm}
\usepackage{mathtools}
\usepackage{float}
\usepackage{placeins}
\usepackage{url}
\usetikzlibrary{arrows,decorations.markings}

\usepackage{amsthm}
\theoremstyle{remark}
\newtheorem{defn}{Definition}
\newtheorem{imp}{Implication}
\newtheorem{hypotest}{Hypothesis test}

\def\QED{\mbox{\rule[0pt]{1.5ex}{1.5ex}}}
\def\endproof{\hspace*{\fill}~\QED\par\endtrivlist\unskip}

\def\bfv{{{\mathbf v}}}

\def\bfx{{{\mathbf x}}}

\def\R{\ensuremath{{\mathbb R}}\xspace}

\def\X{\ensuremath{{\mathcal X}}\xspace}

\def\EMD{{\textit{EMD}}}
\def\SSIM{{\textit{SSIM}}}

\title{Should Adversarial Attacks Use Pixel $p$-Norm?}

\author{%
  Ayon Sen, Xiaojin Zhu, Liam Marshall, and Robert Nowak\\
  University of Wisconsin-Madison\\
  \texttt{\{asen6, limarshall, rdnowak\}@wisc.edu, jerryzhu@cs.wisc.edu} 
}

\begin{document}

\maketitle

\begin{abstract}
Adversarial attacks aim to confound machine learning systems, while remaining virtually imperceptible to humans. Attacks on image classification systems are typically gauged in terms of $p$-norm distortions in the pixel feature space. We perform a behavioral study, demonstrating that the pixel $p$-norm for any $0\le p \le \infty$, and several alternative measures including earth mover's distance, structural similarity index, and deep net embedding, do not fit human perception.  Our result has the potential to improve the understanding of adversarial attack and defense strategies.
\end{abstract}

\section{Introduction}
Adversarial (test-time) attacks perturb an input item $\bfx_0$ slightly, forming $\bfx$ such that (1) $\bfx$ is classified differently than $\bfx_0$; (2) the change from $\bfx_0$ to $\bfx$ is small. 
The oft-quoted reason for (2) is to make the attack hard to detect~\cite{szegedy2013intriguing,goodfellow6572explaining,moosavi2017universal,carlini2017adversarial}.
This assumes an ``inspector'', who detects suspicious items before sending them to the classifier~\cite{salamati2019perception}. 
Our paper focuses on (2) in the context of image classification attacks where the inspector is a human.
We ask the question: are current measures of ``small change'' adequate to characterize visual detection by a human inspector? Answering this question is directly relevant to the efficacy of adversarial learning research.

\begin{figure}[h]
\begin{minipage}[b]{0.45\linewidth}
\centering
\includegraphics[width=0.9\textwidth]{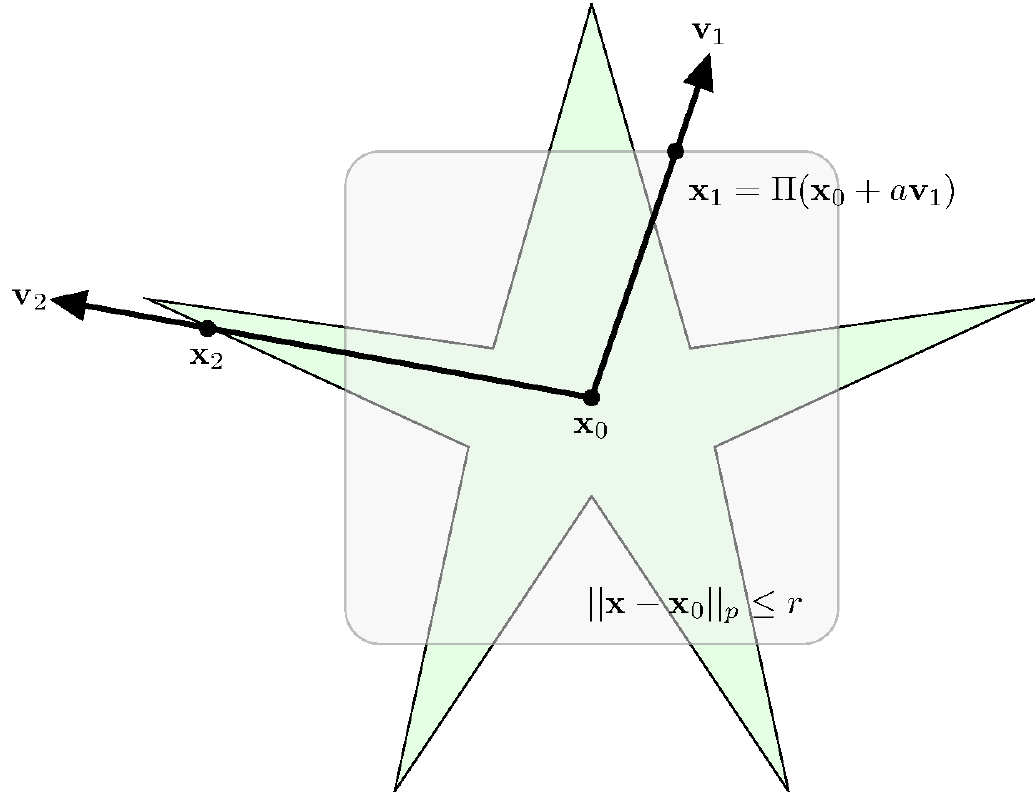}
\caption{Schematic diagram of mismatch between human perception and pixel $p$-norm} 
\label{fig:star}
\end{minipage}
\hspace{0.5cm}
\begin{minipage}[b]{0.45\linewidth}
\centering
\includegraphics[width=\textwidth]{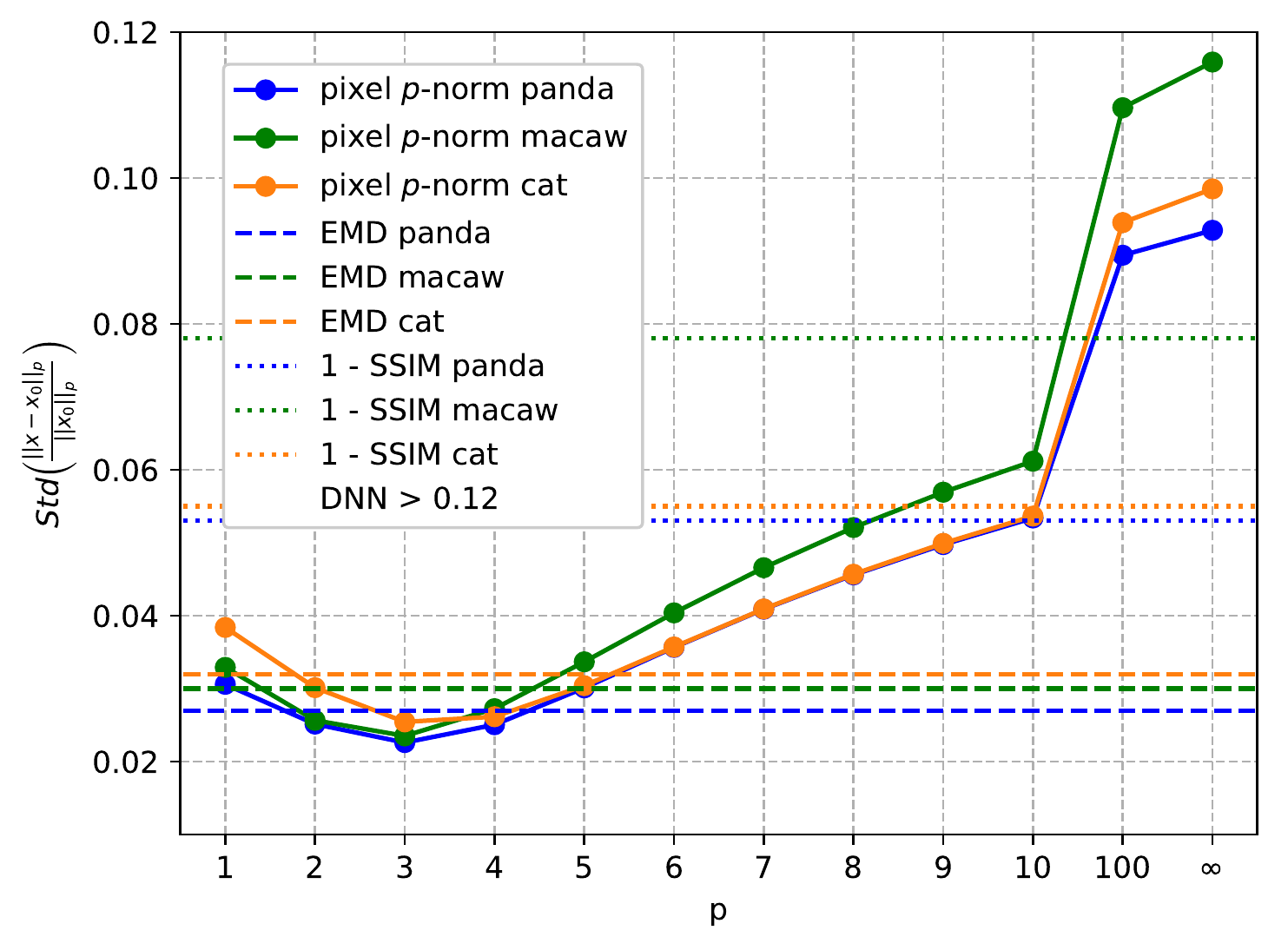}
\caption{Variability of fit to human data, lower is better.  See section~\ref{sec:approx} for details and discussion}
\label{fig:var}
\end{minipage}
\end{figure}

We start with a popular measure: the pixel $p$-norm. Perturbations lying inside the norm ball $\{\bfx: \|\bfx-\bfx_0\|_p \le r\}$ (see Figure~\ref{fig:star}) are assumed to be imperceptible. 
However, even with the optimal norm $p$ and radius $r$, there can be a mismatch between what an average human perceives as small changes to $\bfx_0$ (schematic green area, which may not be in the $p$-norm ball family) and the pixel $p$-norm ball (gray).
Without knowledge of human perceptual behavior, adversarial machine learning can't accurately attack \emph{or} defend: adversaries cannot predict which attacks will succeed, and defenders don't know what is most important to look for. For an adversary, false acceptances like $\bfx_1$ produce futile attacks because the human inspector readily detects the attack, while false rejections like $\bfx_2$ lead to a false sense of security: one may presume the attack at $\bfx_2$ can never escape detection due to its large $p$-norm, while in reality the attack will pass the human inspector.
The need for human perception knowledge is urgent:
Out of 32 papers we surveyed, 27 papers (each with over $100$ citations) used pixel $p$-norms in attacks.
Among these 27, 20\% assumed $p$-norms are a good match to human perception without providing evidence;
50\% used them because other papers did; 
and the rest used them without justification.

Such knowledge requires multiple interdisciplinary studies in adversarial machine learning and cognitive science.
The seminal work by Sharif \textit{et al.} performed a behavioral study on adversarial attack and human perception~\cite{Sharif2018Bright}.
They showed that humans may categorize two perturbed thumbnails -- of the same pixel $p$-norm (for $p=0, 2, \infty$) distance to the original thumbnail -- differently.
While valuable, their conclusions are limited due to the study design:
they only tested pixel $0$-, $2$-, $\infty$-norms but not other $p$-norms or measures.  
Their test also required knowledge of the radius $r$, and depended on humans (mis)-categorizing a low resolution thumbnail (MNIST~\cite{lecun1998mnist}, CIFAR10~\cite{krizhevsky2009learning}), which does not reflect humans' ability to notice small changes in a normal-sized image well before humans' categorization on that image changes.

Our work significantly extends and complements~\cite{Sharif2018Bright}, and addresses all these issues:
Our design enables us to test all pixel $p$-norms, earth mover's distance, structural similarity (SSIM), and deep neural network representation.
It is also agnostic to the true value of $r$ by using the notion of human just-noticeable-difference.
We test humans in small image-change regimes that better match what a human inspector typically faces in an adversarial setting.
\textbf{Our main results caution against the use of pixel $p$-norms, earth mover's distance, structural similarity, or deep neural network representation to define ``small changes'' in adversarial attacks.}
In addition, we give quality of approximation for different measures.
For instance, we experimentally determined (see Figure~\ref{fig:var}) that pixel 3-norm is the best approximation to human data among the measures studied, see section~\ref{sec:approx} for details.
Our results have the potential to improve the understanding of adversarial attack and defense strategies.

We also mention some limitations of our work.
We cannot directly answer ``what is the correct measure'', because computationally modeling human visual perception is still an open question in psychology~\cite{simons2005change,rensink2002change,wolfe2010visual,itti2001computational}.
We used a ``show $\bfx_0$ then perturb'' experiment paradigm, while in real applications the human inspector may not have access to $\bfx_0$.
We also limit ourselves to the visual domain.
These topics remain future work.


\section{The pixel $p$-norm central hypothesis and its implications}

Let the pixel feature space be $\X \coloneqq \{0,\ldots,255\}^d$, where $d$ equals the number of pixels times the number of color channels (it is straightforward to generalize to color depth other than 255).
Consider a natural image $\bfx_0 \in \X$ and another image $\bfx$.
The pixel $p$-norm for any $p> 0$ measures the amount of perturbation by
$\|\bfx-\bfx_0\|_p \coloneqq \left( \sum_{i=1}^d |x_i - x_{0,i}|^p \right)^{1/p}$.
We define the 0-norm to be the number of nonzero elements.
To facilitate mathematical exposition in this section, we posit an ``ideal observer'' who has \emph{population median} human perception.
Natural variations in real human observers will be handled in section~\ref{sec:refute}.
The \emph{central hypothesis} of pixel $p$-norm is the following.

\begin{defn}[The Central Hypothesis]
	$\exists p^* \ge 0$, $\forall \bfx_0, \exists$threshold $r(\bfx_0)$, such that the ideal observer perceives any $\bfx$ the same as $\bfx_0$ if $\|\bfx-\bfx_0\|_{p^*} < r(\bfx_0)$, and the ideal observer notices the difference if $\|\bfx-\bfx_0\|_{p^*} \ge r(\bfx_0)$.
\end{defn}

The threshold $r(\bfx_0)$ is known as the ``Just Noticeable Difference'' (JND) in experimental psychology~\cite{fechner1966elements,zhang2008just}. 
We further define the set of Just-Noticeably-Different images with respect to $\bfx_0$ under the central hypothesis: 
$J(\bfx_0) \coloneqq\{\bfx \subset \X: \|\bfx-\bfx_0\|_{p^*} = r(\bfx_0)\}$.
In other words, $J(\bfx_0)$ is the shell of the norm ball centered at $\bfx_0$ with radius $r(\bfx_0)$. 
A main task of the present paper is to test the central hypothesis.
To this end, we derive a number of testable implications of the central hypothesis.
These implications will be tested through human behavioral experiments in later sections.
The first implication follows trivially from the definition of $J(\bfx_0)$.
It states that any Just-Noticeably-Different images of an $\bfx_0$ has the same $p^*$-norm (note: does not require knowledge of $r(\bfx_0)$):
\begin{imp}
	\label{imp:equalnorm}
	Suppose $p^*$ is the correct norm for the central hypothesis.  Then $\forall \bfx_0, \forall \bfx_1, \bfx_2 \in J(\bfx_0), \|\bfx_1-\bfx_0\|_{p^*} = \|\bfx_2-\bfx_0\|_{p^*}$.
\end{imp}

The second implication is more powerful in the sense that it can be tested without knowing the true parameter $p^*$ or $r(\bfx_0)$.
To do so, we utilize special perturbed $\bfx$ as follows.
As indicated in Figure~\ref{fig:star}, we consider $\bfx$ generated along the ray defined by a \textbf{perturbation direction} $\bfv \in \R^d$ with a \textbf{perturbation scale} $a > 0$:
$\bfx = \Pi\left( \bfx_0 + a \bfv\right)$.
Here $\Pi$ is the projection onto $\X$; namely, clipping values to $[0, 255]$ and rounding to integers.
Note that as $a$ increases, the perturbation becomes stronger.
The perturbation direction $\bfv$ is important: in our experiments some directions are generated by popular adversarial attacks in the literature, while others are designed to facilitate statistical tests.
Specifically, we define \textbf{$\pm 1$-perturbation directions} as any $\bfv \in \R^d$ with the following two properties:
	(i) Its support (nonzero elements) has cardinality $s>0$; in many cases $\bfv$ will be sparse with $s \ll d$; 
	(ii) the nonzero elements $v_i$ are either 1 or -1 depending on the value of the corresponding element $x_{0,i}$ in $\bfx_0$:
	$v_i=1$ if $x_{0,i} < 128$, and -1 otherwise.
For $\pm 1$-perturbations $\bfv$ and integer $a \in \{1,\ldots,128\}$ it is easy to see that the projection $\Pi$ is not needed: 
$\bfx=\Pi\left( \bfx_0 + a \bfv\right) = \bfx_0 + a \bfv$.
This allows for convenient experiment design.
More importantly, for such $\pm 1$-perturbed images any pixel $p$-norm has a simple form:
$\forall p: \|\bfx - \bfx_0\|_p = \left( \sum_{\bfv_i \neq 0} |a v_i|^p \right)^{1/p} = a s^{1/p}$.
Implication~\ref{imp:same_s} states that two just-noticeable perturbed images with the same perturbation sparsity $s$ should have the same perturbation scale $a$.
Importantly, it can be tested without knowing $p^*$ or $r(\bfx_0)$.
If it fails then no pixel $p$-norm is appropriate to model human perceptions of just-noticeable-difference.
\begin{imp}
	\label{imp:same_s}
	$\forall p>0$, $\forall \bfx_0$, $\forall \pm 1$-perturbation directions $\bfv_1, \bfv_2$ with the same sparsity $s$,
	suppose $\exists a_1, a_2 \in \{1, \dots, 128\}$ such that
	$\bfx_1 = \bfx_0 + a_1 \bfv_1 \in J(\bfx_0)$ and 
	$\bfx_2 = \bfx_0 + a_2 \bfv_2 \in J(\bfx_0)$. 
	Then $a_1 = a_2$.
\end{imp}



\section{Behavioral experiment design}
We conducted a human behavioral experiment under Institutional Review Board (IRB) approval.
We release all behavioral data, and the code that produces the plots and statistical tests in this paper, to the public for reproducibility and further research at \url{http://www.cs.wisc.edu/~jerryzhu/pub/advMLnorm/}.
The figures below are best viewed by zooming in to replicate the participant experience.

\begin{figure}[t]
\begin{center}
(a) 
			\includegraphics[width=0.15\textwidth]{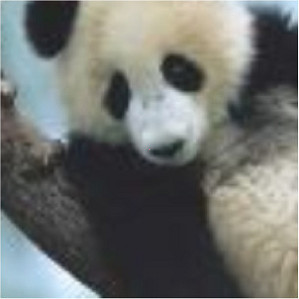}
			\includegraphics[width=0.15\textwidth]{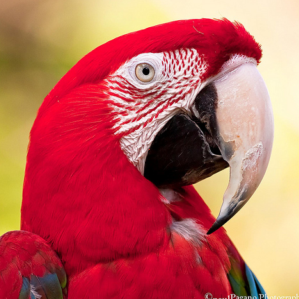}
			\includegraphics[width=0.15\textwidth]{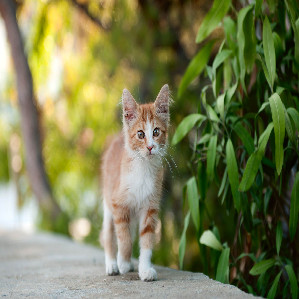}
(b)
			\includegraphics[width=0.15\textwidth]{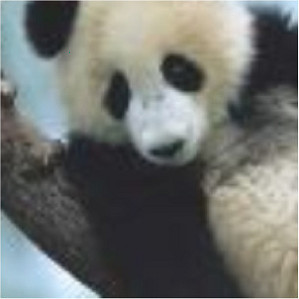}
			\includegraphics[width=0.15\textwidth]{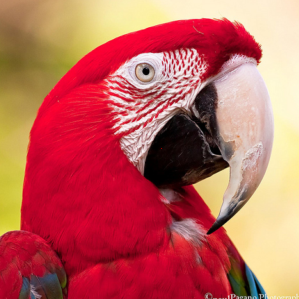}
			\includegraphics[width=0.15\textwidth]{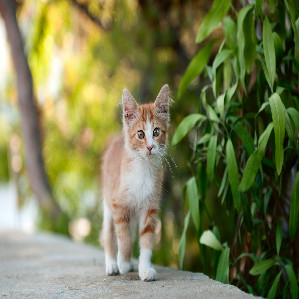} 
\\
(c)
			\includegraphics[width=0.15\textwidth]{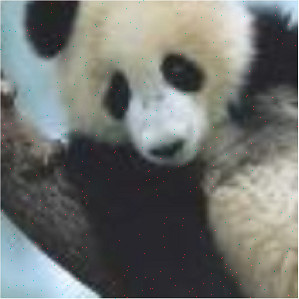}
			\includegraphics[width=0.15\textwidth]{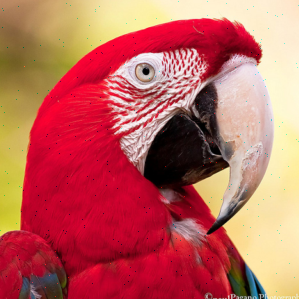}
			\includegraphics[width=0.15\textwidth]{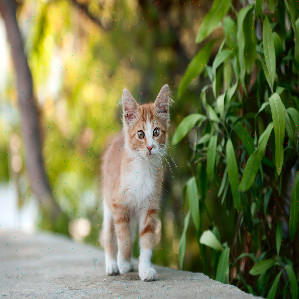}
(d)
			\includegraphics[width=0.15\textwidth]{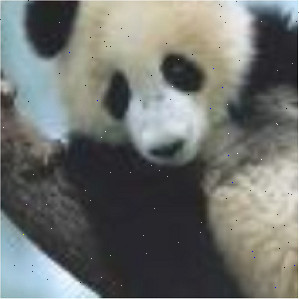}
			\includegraphics[width=0.15\textwidth]{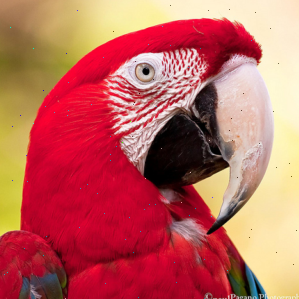}
			\includegraphics[width=0.15\textwidth]{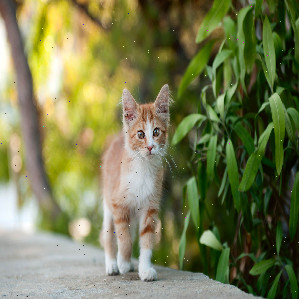}
\\
(e)
			\includegraphics[width=0.15\textwidth]{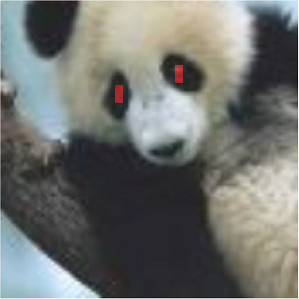}
			\includegraphics[width=0.15\textwidth]{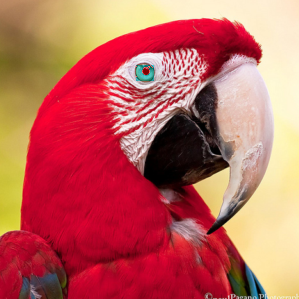}
			\includegraphics[width=0.15\textwidth]{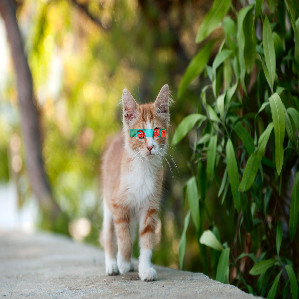}
(f)
			\includegraphics[width=0.15\textwidth]{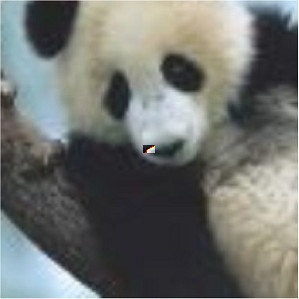}
			\includegraphics[width=0.15\textwidth]{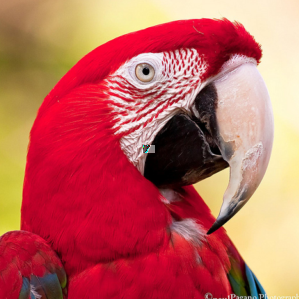}
			\includegraphics[width=0.15\textwidth]{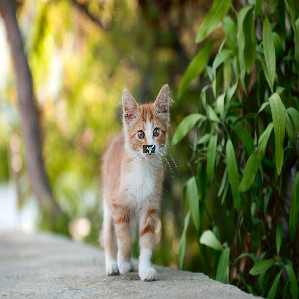}
\\
(g)
			\includegraphics[width=0.15\textwidth]{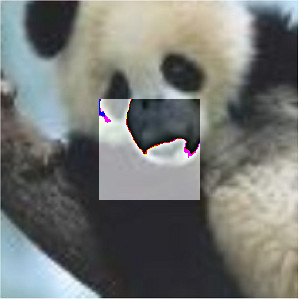}
			\includegraphics[width=0.15\textwidth]{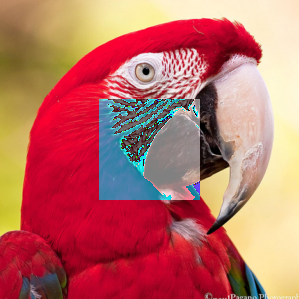}
			\includegraphics[width=0.15\textwidth]{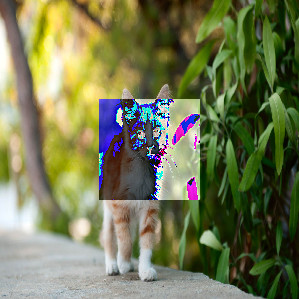}
(h)
			\includegraphics[width=0.15\textwidth]{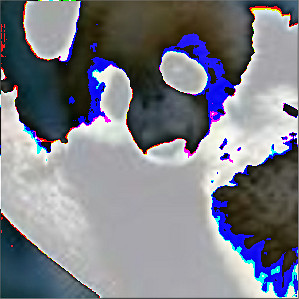}
			\includegraphics[width=0.15\textwidth]{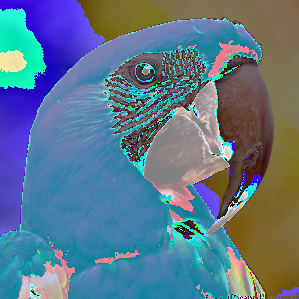}
			\includegraphics[width=0.15\textwidth]{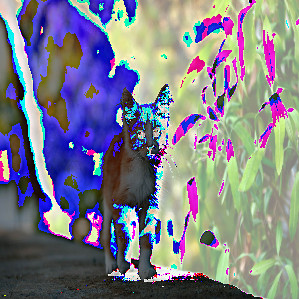}
\\
(i)
\includegraphics[width=0.15\textwidth]{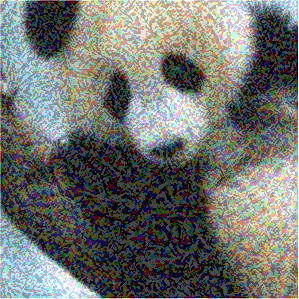}
\includegraphics[width=0.15\textwidth]{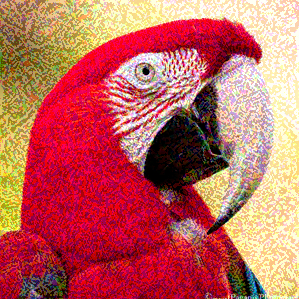}
\includegraphics[width=0.15\textwidth]{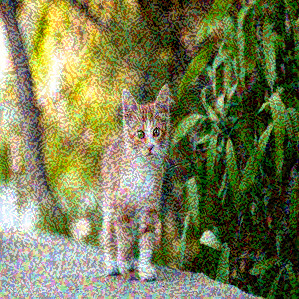}
(j)
\includegraphics[width=0.15\textwidth]{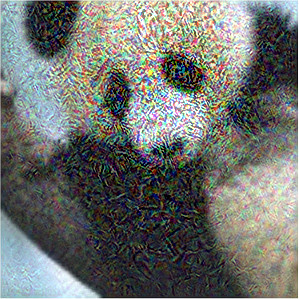}
\includegraphics[width=0.15\textwidth]{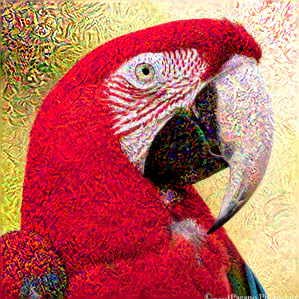}
\includegraphics[width=0.15\textwidth]{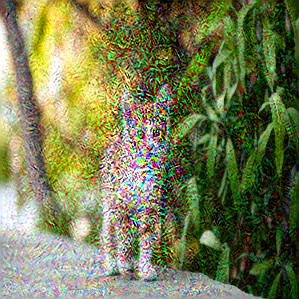}
\end{center}
	\caption{All 10 perturbation directions $\bfv$ with severe perturbation scale $a=128$.
(a) S\_Red\_Box: the red channel of the center pixel.
(b) S\_Red\_Dot: a randomly selected red channel. 
(c) M\_Red\_Dot: 288 randomly selected red channels.
(d) M\_RGB\_Dot: all three color channels of 96 randomly selected pixels ($s= 3\times 96 = 288$). 
(e) M\_Red\_Eye: 288 red channels around the eyes of the animals. 
(f) M\_RGB\_Box: all colors of a centered $8 \times 12$ rectangle.
(g) L\_RGB\_Box: all colors of a centered $101 \times 101$ rectangle. 
(h) X\_RGB\_Box: all dimensions. 
(i) FGSM.
(j) PGD.
}
	\label{fig:perturbed_images}
\end{figure}

\textbf{Center images $\bfx_0$ and perturbation directions $\bfv$}:
We chose three natural images (from the Imagenet dataset~\cite{deng2009imagenet}) popular in adversarial research: a panda~\cite{goodfellow6572explaining}, a macaw~\cite{moosavi2016deepfool} and a cat~\cite{athalye2018obfuscated} as $\bfx_0$ in our experiment.
We resized the images to $299\times299$ to match the input dimension of the Inception V3 image classification network~\cite{szegedy2016rethinking}.
For each natural image $\bfx_0$ we considered 10 perturbation directions $\bfv$, see Figure~\ref{fig:perturbed_images}.
Eight are specially crafted $\pm 1$-perturbation directions varying in three attributes, and further explained in the caption of Figure~\ref{fig:perturbed_images}:
\begin{center}
\begin{small}
	\begin{tabular}{l|l|l}
		\multicolumn{1}{c|}{\textbf{\# Dimensions Changed (s)}} & \multicolumn{1}{c|}{\textbf{Color Channels Affected}} & \multicolumn{1}{c}{\textbf{Shape of Perturbed Pixels}} \\\hline
		S = 1, M = 288 & Red = only the red channel of a pixel & Box = a centered rectangle \\
		L = 30603, X = 268203 & RGB = all three channels of a pixel & Dot = scattered random dots \\
		(mnemonic: garment size) & & Eye = on the eye of the animal \\
	\end{tabular}
\end{small}
\end{center}

The remaining two perturbation directions are adversarial directions. 
We used Fast Gradient Sign Method (FGSM)~\cite{goodfellow6572explaining} and Projected Gradient Descent (PGD)~\cite{madry2017towards} to generate two adversarial images $\bfx^{FGSM}, \bfx^{PGD}$ for each $\bfx_0$, with Inception V3 as the victim network.
All attack parameters are set as suggested in the methods' respective papers.
PGD is a directed attack and requires a target label; we choose gibbon (on panda) and guacamole (on cat) following the papers, and cleaver (on macaw) arbitrarily.
We then define the adversarial perturbation directions by 
$\bfv^{FGSM} = 127.5 (\bfx^{FGSM} - \bfx_0)/{\Vert \bfx^{FGSM} - \bfx_0 \Vert_2}$ 
and
$\bfv^{PGD} = 127.5 (\bfx^{PGD} - \bfx_0)/{\Vert \bfx^{PGD} - \bfx_0 \Vert_2}$. 
We use the factor $127.5$ based on a pilot study to ensure that changes between consecutive images in the adversarial perturbation directions are not too small or too big.
%
%
%
%

\begin{figure}[htb]
	\centering
	\includegraphics[width=\textwidth]{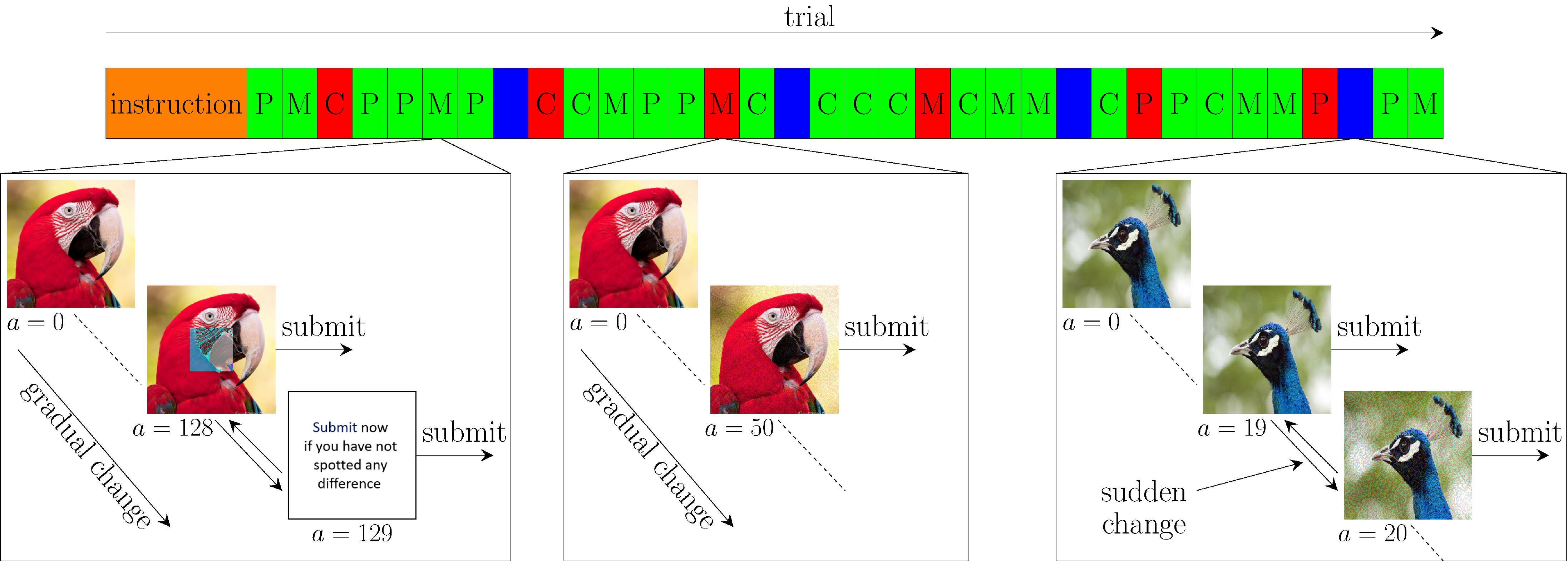}
	\caption{Experiment procedure.
The green, red and blue cells denote $\pm 1$-perturbation, adversarial, and guard trials, respectively. 
The letters P, M and C denote the panda, macaw and cat $\bfx_0$, respectively.
}
\label{fig:temporal}
\end{figure}

\textbf{Experimental procedure}:
See Figure~\ref{fig:temporal}.
Each participant was first presented with instructions and then completed a sequence of $34$ trials, of which 30 were $\pm 1$-perturbation or adversarial trials, and 4 were guard trials.
The order of these trials was randomized then fixed (see figure).
During each trial the participants were presented with an image $\bfx_0$.
They were instructed to increase (decrease) perturbations to this image by using right / left arrow keys or buttons.
Moving right (left) incremented (decremented) $a$ by 1, and the subject was then presented with the new perturbed image $\bfx=\Pi\left( \bfx_0 + a \bfv\right)$.
We did not divulge the nature of the perturbations $\bfv$ beforehand, nor the current perturbation scale $a$ the participant had added to $\bfx_0$ at any step of the trial.
\textbf{
The participants were instructed to submit the perturbed image $\bfx$ when they think it became just noticeably different from the original image $\bfx_0$}.
The participants had to hold $\bfx_0$ in memory, though they could also go all the way left back to see $\bfx_0$ again. 
We hosted the experiment using the NEXT platform~\cite{jamieson2015next,sievert2017next}.

In a $\pm 1$-perturbation trial, the perturbation direction $\bfv$ is one of the eight $\pm 1$-perturbations.
We allowed the participants to vary $a$ within $\{0,1,\ldots,128\}$ to avoid value cropping.  If a participant was not able to detect any change even after $a=128$, then they were encouraged to ``give up'' (see figure).

\begin{figure}[ht]
\begin{center}
\includegraphics[width=0.32\textwidth]{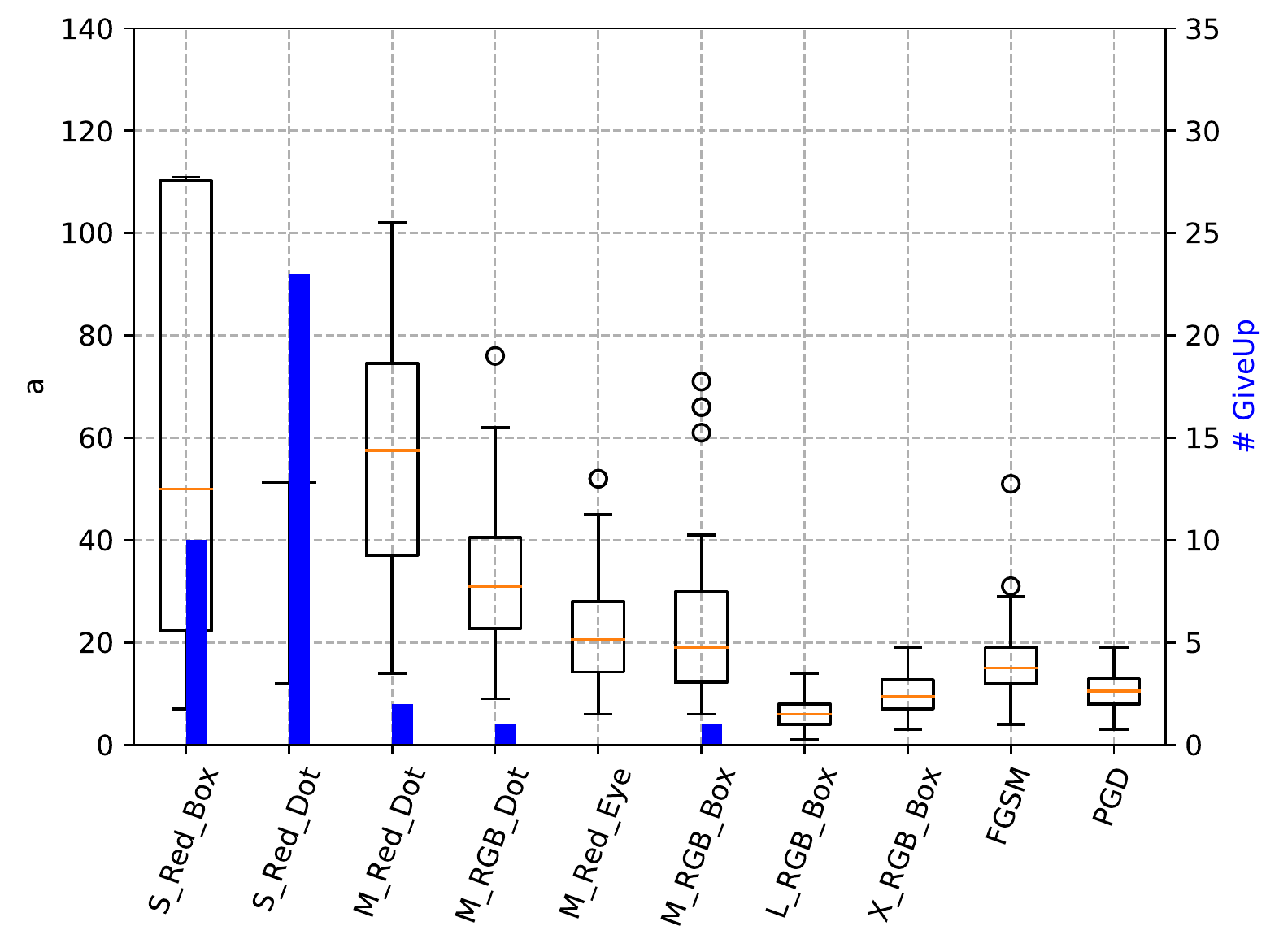}
\includegraphics[width=0.32\textwidth]{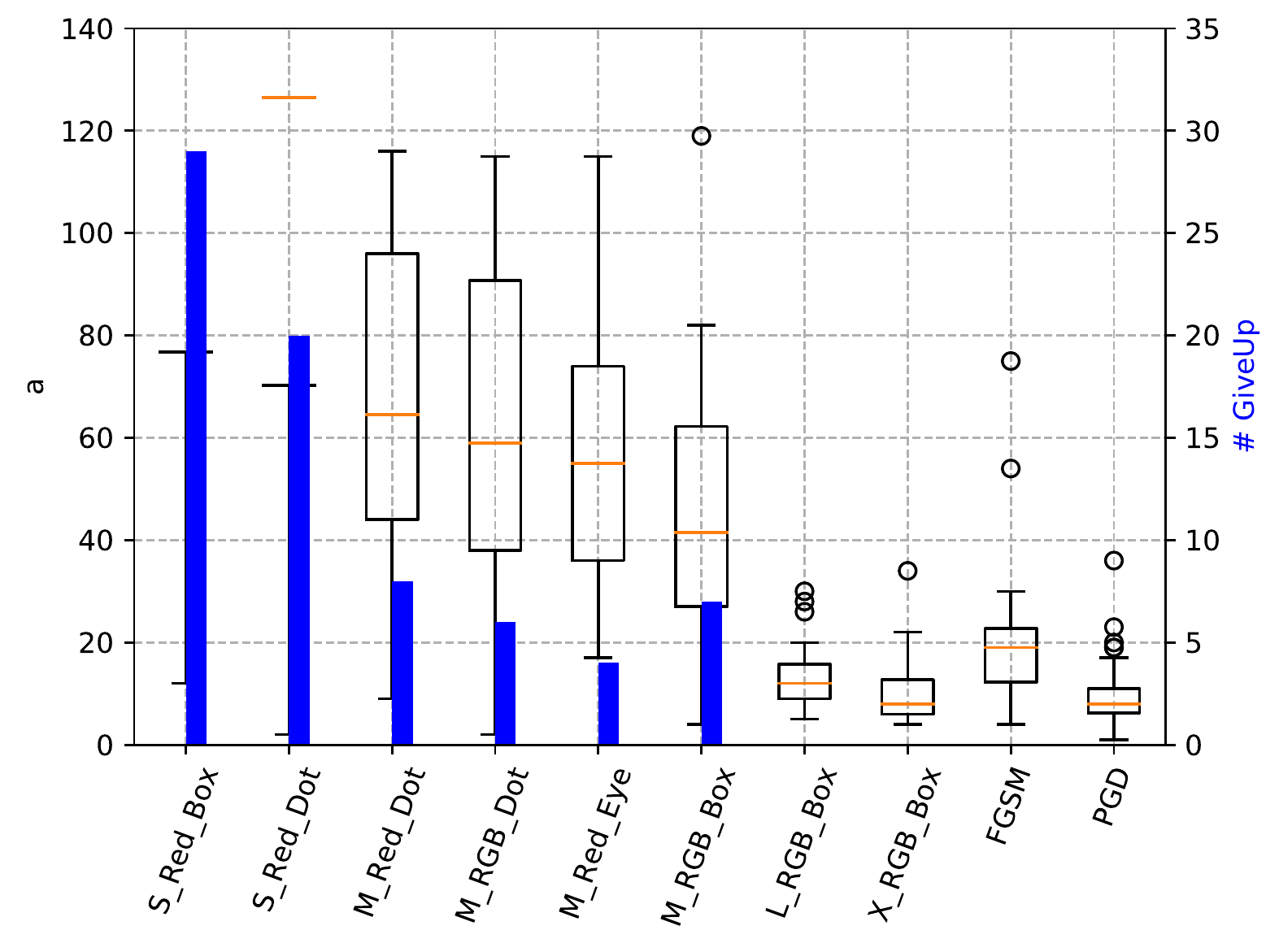}
\includegraphics[width=0.32\textwidth]{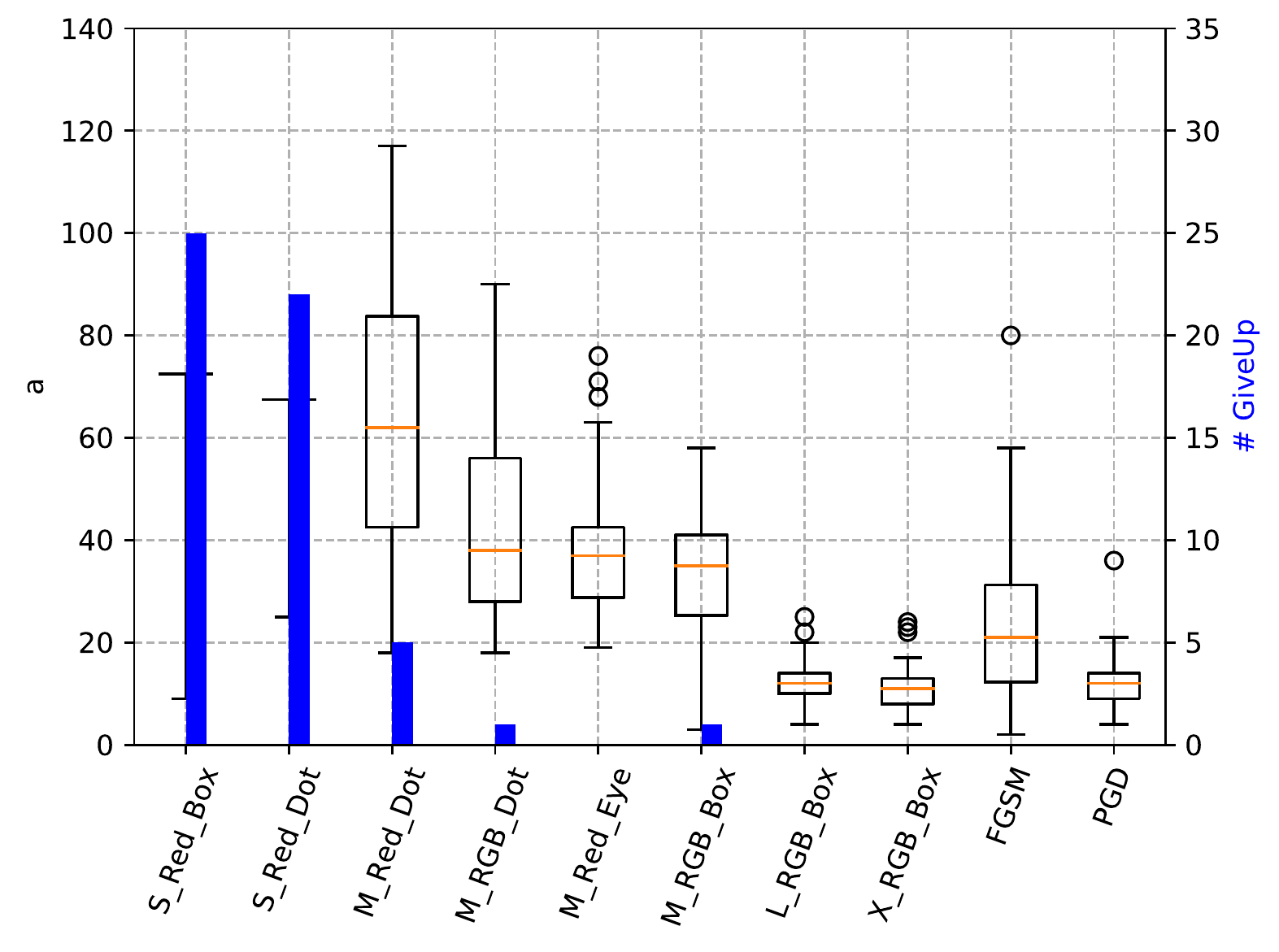}
\caption{Summary of data for $\bfx_0=$ panda, macaw, cat, respectively}
\label{fig:boxplot_a}
\end{center}
\end{figure}

In an adversarial trial, the perturbation direction is $\bfv^{FGSM}$ or $\bfv^{PGD}$.  We allowed the participants to increment $a$ indefinitely, though no one went beyond $a=80$, see Figure~\ref{fig:boxplot_a}.
	
The guard trials were designed to filter out participates who clicked through the experiment without performing the task.
In a guard trial, we showed a novel fixed natural image (not panda, macaw or cat) for $a < 20$.
Then for $a \ge 20$, a highly noisy version of that image 
is displayed.
An attentive participant should readily notice this sudden change at $a=20$ and submit it.
In our main analyses, we disregarded the guard trials.

%

\textbf{Participants and data inclusion criterion}:
We enrolled 68 participants using Amazon Mechanical Turk~\cite{buhrmester2011amazon} master workers.
A master worker is a person who has consistently displayed a high degree of success in performing a wide range of tasks. 
All participants used a desktop, laptop or a tablet device; none used a mobile device where the screen would be too small. 
On average the participants took $33$ minutes to finish the experiment.
Each participant was paid $\$5$. 
As mentioned before, we use guard trials to identify inattentive participants.
While the change happens at exactly $a=20$ in a guard trial, our data indicates a natural spread in participant submissions around 20 with sharp decays.  We speculate that the spread was due to keyboard / mouse auto repeat.
We set a range for an acceptable guard trial if a participant submitted $a \in \{18, 19, 20, 21, 22\}$.
A participant is deemed inattentive if any one of the four guard trials was outside the acceptable range.
Only $n=42$ out of 68 participants survived this stringent inclusion condition.
All our analyses below are on these 42 participants.

\textbf{To summarize the data}: on each combination of natural image $\bfx_0$ and perturbation direction $\bfv$, the $n$ participants gave us their individual perturbation scale $a^{(1)}, \ldots, a^{(n)}$.
That is, the image $\bfx = \Pi(\bfx_0+a^{(j)} \bfv)$ is the one participant $j$ thinks has just-noticeable-difference to $\bfx_0$.   We will call these human JND images.
We present box plots of the data in Figure~\ref{fig:boxplot_a}. 
The perturbation directions $\bfv$ are indicated on the x-axis.
The box plots (left y-axis) show the median, quartiles, and outliers of the participants' perturbation scale $a$.

Because our participants can sometimes choose to ``give up'' if they did not notice a change, we have \emph{right censored data} on $a$.
All we know from a give-up trial is that $a \ge 129$, but not what larger $a$ value will cause  the participant to noticed a difference.
In Figure~\ref{fig:boxplot_a} the blue bars (right y-axis) show the number of participants who chose to ``give up''. 
Not surprisingly, many participants failed to notice a difference along the S\_Red\_Box and S\_Red\_Dot perturbation directions.
Because of the presence of censored data, in later sections we often employ the Kolmogorov-Smirnov test which is a nonparametric test of distribution that can incorporate the censored data.
{There are 9 tests including the appendix. To achieve a paper-wide significance level of e.g. $\alpha=0.01$, we perform Bonferroni correction for multiple tests leading to individual test level $\alpha/9$.}

\section{Pixel $p$-norms do not match human perception}
\label{sec:refute}

\subsection{Humans probably do not use pixel 0-norm, 1-norm, 2-norm, or $\infty$-norm}
Let us start by assuming humans use pixel 1-norm, i.e. $p^*=1$.
Implication~\ref{imp:equalnorm} suggests the following procedure: for all original images $\bfx_0$, for all perturbation directions $\bfv_1, \bfv_2$, perturb $\bfx_0$ along these two directions separately until the images each become just noticeable to the ideal observer.   
Denote $t_{1} \coloneqq \|\bfx_1 - \bfx_0\|_1$ and $t_{2} \coloneqq \|\bfx_2 - \bfx_0\|_1$ on the two resulting images $\bfx_1, \bfx_2$.
Then we have $t_1=t_2$.
Conversely, if the equality does not hold on even one triple ($\bfx_0$, $\bfv_1, \bfv_2$), then implication~\ref{imp:equalnorm} with $p^*=1$, and consequently the central hypothesis with $p^*=1$, will be refuted. 

\begin{figure}[h]
	\begin{center}
		\begin{tabular}{ccc}
			$p=1,\ \bfx_0=\text{panda}$ & $p=2,\ \bfx_0=\text{macaw}$ & $p=\infty,\ \bfx_0=\text{cat}$ \\
			\includegraphics[width=0.26\textwidth]{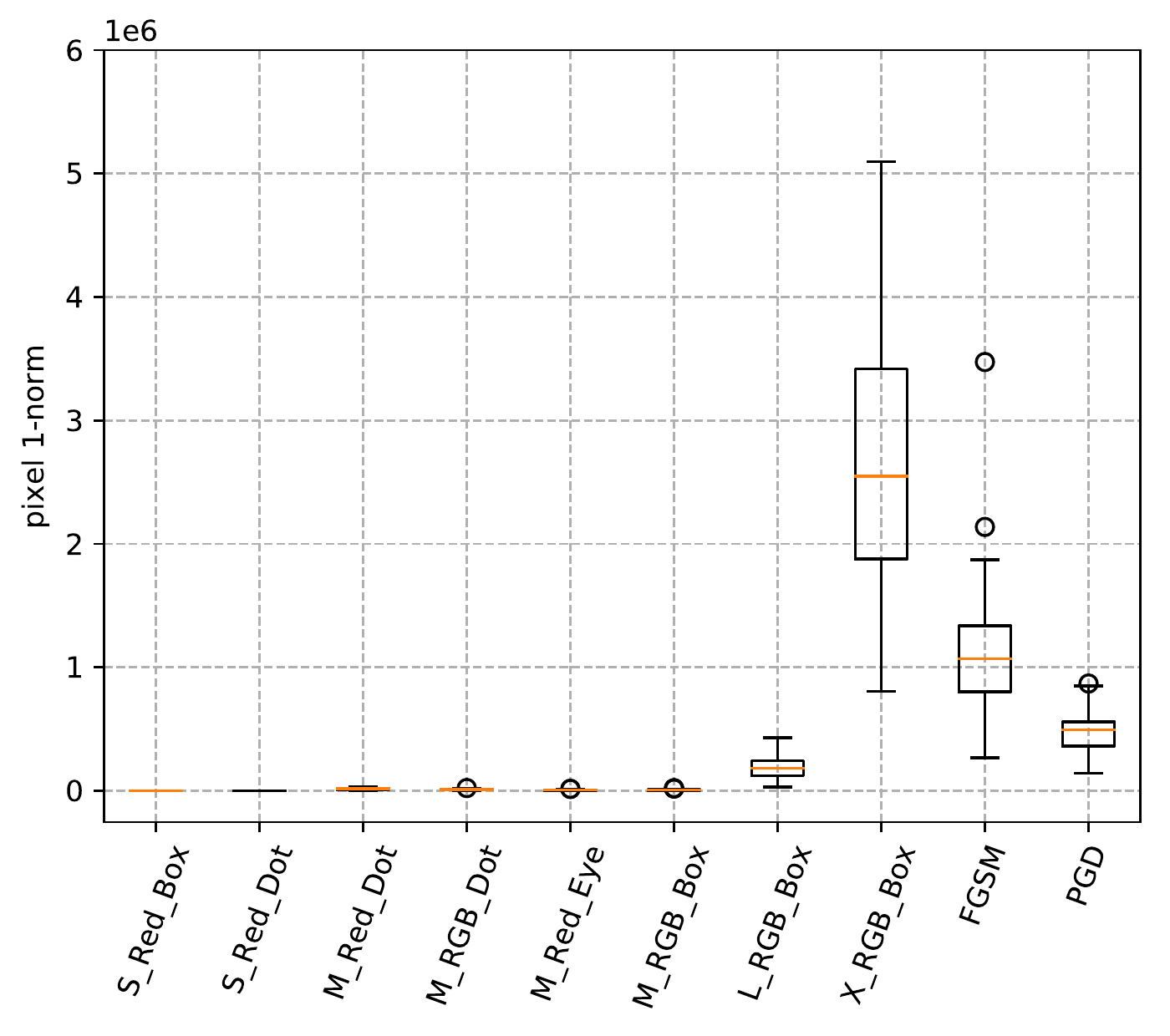} &
			\includegraphics[width=0.26\textwidth]{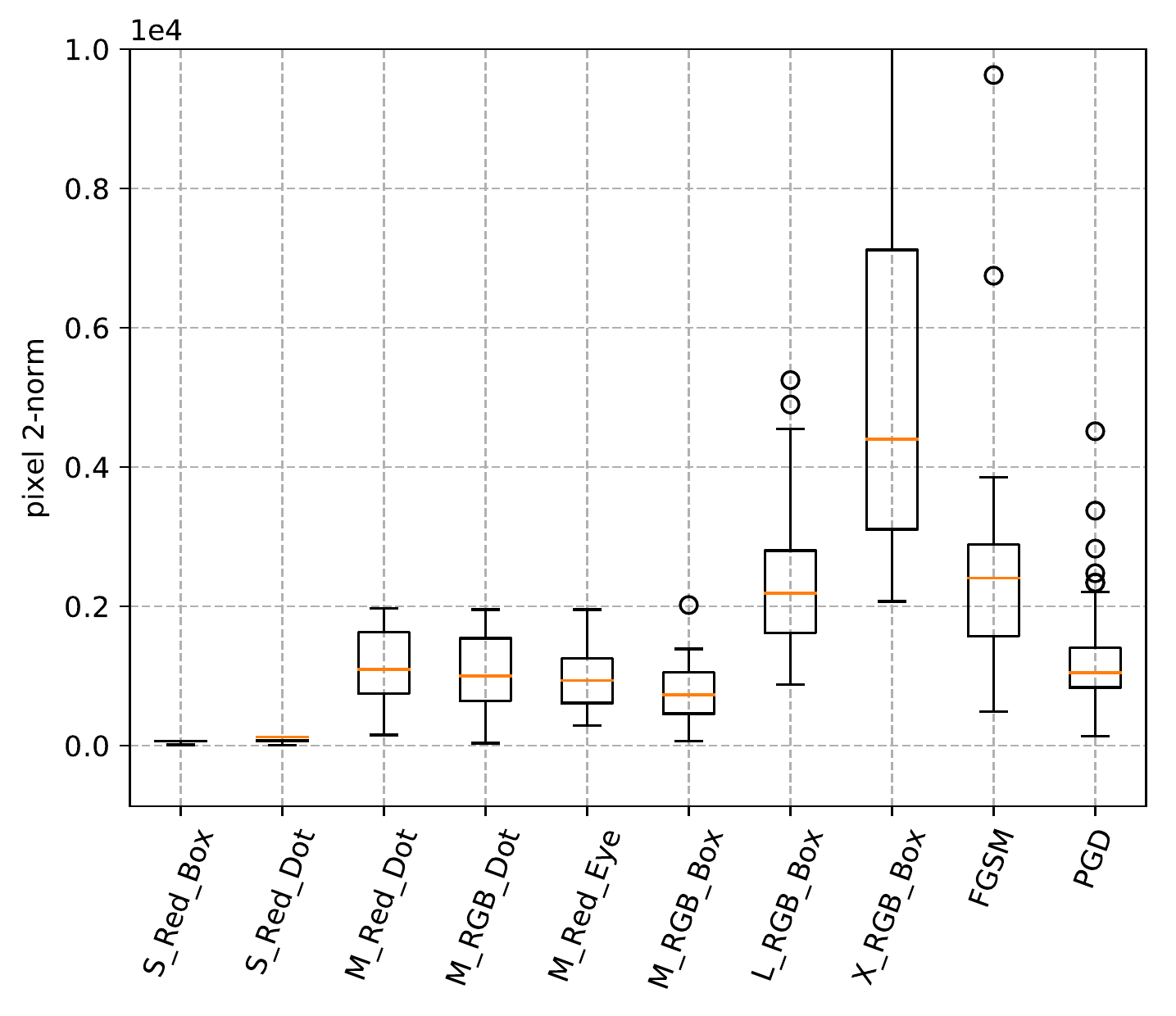} &
			\includegraphics[width=0.26\textwidth]{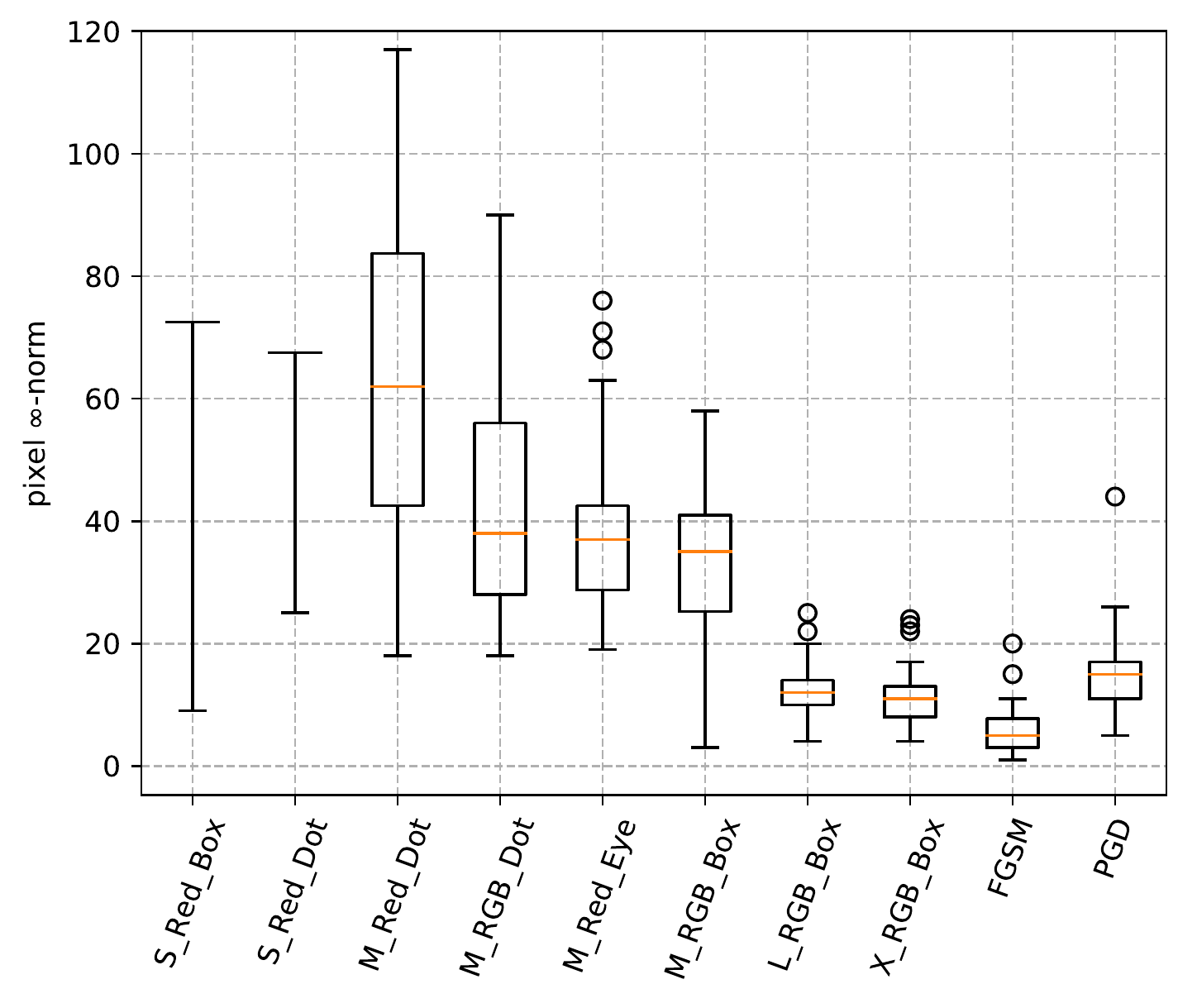} 
		\end{tabular}
	\end{center}
	\caption{Participant JND $\bfx$'s pixel norm $\|\bfx-\bfx_0\|_p$.  If the central hypothesis were true, one expects a plot to have similar medians (orange lines).}
	\label{fig:Lp_excerpt}
\end{figure}

Of course, we do not have the ideal observer.
Instead, we have $n$ participants from the population.
Starting from $\bfx_0$ along perturbation direction $\bfv_1$, the $j$th participant identifies their own just-noticeably-different image $\bfx_1^{(j)}$.
Under $p^*=1$ this produces a number $t_{1j} \coloneqq \|\bfx_1^{(j)} - \bfx_0\|_1$.
The numbers from all participants form a sample $\{t_{11}, \ldots, t_{1n}\}$ (there can be identical values).
Similarly, denote the sample for direction $\bfv_2$ by $\{t_{21}, \ldots, t_{2n}\}$.
Figure~\ref{fig:Lp_excerpt}(left) shows a box plot for $\bfx_0=\text{panda}$. 
If implication~\ref{imp:equalnorm} with $p^*=1$ were true, the medians (orange lines) would be at about the same height within the plot.
Qualitatively this is not the case:
the median for $\bfv_1=\text{FGSM}$ is $t=1068581$ but the median for $\bfv_2=\text{M-RGB-Dot}$ is merely $t=8928$.
We perform a statistical test.
\begin{hypotest}
	The null hypothesis $H_0$ is: $\|\bfx_1^{(j)} - \bfx_0\|_1$ and $\|\bfx_2^{(j)} - \bfx_0\|_1$ have the same distribution, where $\bfx_0=\text{panda}$, $\bfv_1=\text{FGSM}$ and $\bfv_2=\text{M-RGB-Dot}$. 
	A two-sample Kolmogorov-Smirnov (KS) test on our data ($n=42$) yields a p-value $6.4\times 10^{-19}$, rejecting $H_0$.
\end{hypotest}
Exactly the same reasoning applies if we assume humans use pixel 2-norm or $\infty$-norm.
Figure~\ref{fig:Lp_excerpt}(center) shows 2-norm on $\bfx_0=\text{macaw}$, where $\bfv_1=\text{PGD}$ has median $t \approx 1049$ but $\bfv_2=\text{X-RGB-Box}$ has median $t\approx 4402$;
(right) shows $\infty$-norm on $\bfx_0=\text{cat}$, where $\bfv_1=\text{M-RGB-Box}$ has median $t=35$ but $\bfv_2=\text{L-RGB-Box}$ has median $t=12$.
The full plots are in appendix Figure~\ref{fig:Lp}.
\begin{hypotest}
	$H_0$: $\|\bfx_1^{(j)} - \bfx_0\|_2$ and $\|\bfx_2^{(j)} - \bfx_0\|_2$ have the same distribution, where $\bfx_0=\text{macaw}$, $\bfv_1=\text{PGD}$ and $\bfv_2=\text{X-RGB-Box}$. 
	KS test yields p-value $2.6\times 10^{-16}$, rejecting $H_0$.
\end{hypotest}
\begin{hypotest}
	$H_0$: $\|\bfx_1^{(j)} - \bfx_0\|_\infty$ and $\|\bfx_2^{(j)} - \bfx_0\|_\infty$ have the same distribution, where $\bfx_0=\text{cat}$, $\bfv_1=\text{M-RGB-Box}$ and $\bfv_2=\text{L-RGB-Box}$. 
	KS test yields p-value $1.1\times 10^{-14}$, rejecting $H_0$.
\end{hypotest}

Finally, $p^*=0$ is refuted by noticing in Figure~\ref{fig:boxplot_a} that $\bfx-\bfx_0$ in the M, L, X directions have vastly different 0-norms, yet each direction has its own nonzero scale $a$ that induces human JND.
This contradicts with implication~\ref{imp:equalnorm} with $p^*=0$, which predicts changes are never noticeable below a 0-norm threshold, and always noticeable above it, regardless of $a$. 
Taken together, we have rejected implication~\ref{imp:equalnorm} with $p^*=0, 1, 2$, or $\infty$.
This suggests that humans probably do not use pixel $0$-, $1$-, $2$-, or $\infty$-norm when they judge if a perturbed image is different from its original.

\subsection{Humans probably do not use any pixel $p$-norm}
But what if humans use some other $p^*$-norm in $(0, \infty)$?
Implication~\ref{imp:equalnorm} requires a specific $p^*$ to test, which is not convenient.
Instead, we now test implication~\ref{imp:same_s} whose failure can refute any $p^*$.
We take $\bfx_0=\text{cat}$ and look at the two perturbation directions $\bfv_1=\text{M-Red-Dot}$ and $\bfv_2=\text{M-Red-Eye}$.
These two perturbations have the same sparsity $s=288$.
Therefore, implication~\ref{imp:same_s} predicts that the scales $a_1$, $a_2$ to reach just-noticeable-difference should be the same.
However, the perturbation directions differ in their ``shape of support'': 
M-Red-Dot changes random pixels, while
M-Red-Eye changes pixels of the eye region which presumably humans pay attention to and thus detect earlier.
On perturbation direction $\bfv_1$, our $n$ participants produced scales $\{a_1^{(1)}, \ldots, a_1^{(n)}\}$; similarly, for the other direction $\bfv_2$, they produced $\{a_2^{(1)}, \ldots, a_2^{(n)}\}$.
See Figure~\ref{fig:boxplot_a}(right) for the human behaviors: the median scale is 62 and 37, respectively, as we suspected.
\begin{hypotest}
	$H_0$: Human JND $a_1^{(j)}$ and $a_2^{(j)}$ have the same distribution, where $\bfx_0=\text{cat}$, $\bfv_1=\text{M-Red-Dot}$ and $\bfv_2=\text{M-Red-Eye}$.
	KS test yields p-value $9.5\times 10^{-6}$, rejecting $H_0$.
\end{hypotest}

We report more statistical tests in the appendix that further refute this and other implications.
Taken together, these results indicate that pixel $p$-norms are not a good fit for human behaviors regardless of $p$.
There are probably other perceptual attributes that are important to humans which are unaccounted for by pixel $p$-norms.

\section{Some measures other than pixel $p$-norms}
We seek an alternative distance function (does not need to be a metric) $\rho: \X \times \X \mapsto \R_+$ that matches human perception.  That is, for human JND images in $J(\bfx_0)$, ideally $\rho$ satisfies
$\forall \bfx_1, \bfx_2 \in J(\bfx_0), \rho(\bfx_1, \bfx_0)=\rho(\bfx_2,\bfx_0)$.

\begin{figure}[!h]
\begin{center}
\begin{tabular}{ccc}
1-SSIM, $\bfx_0=\text{panda}$ &  
DNN, $\bfx_0=\text{macaw}$ & 
EMD, $\bfx_0=\text{cat}$ \\
\includegraphics[width=0.26\textwidth]{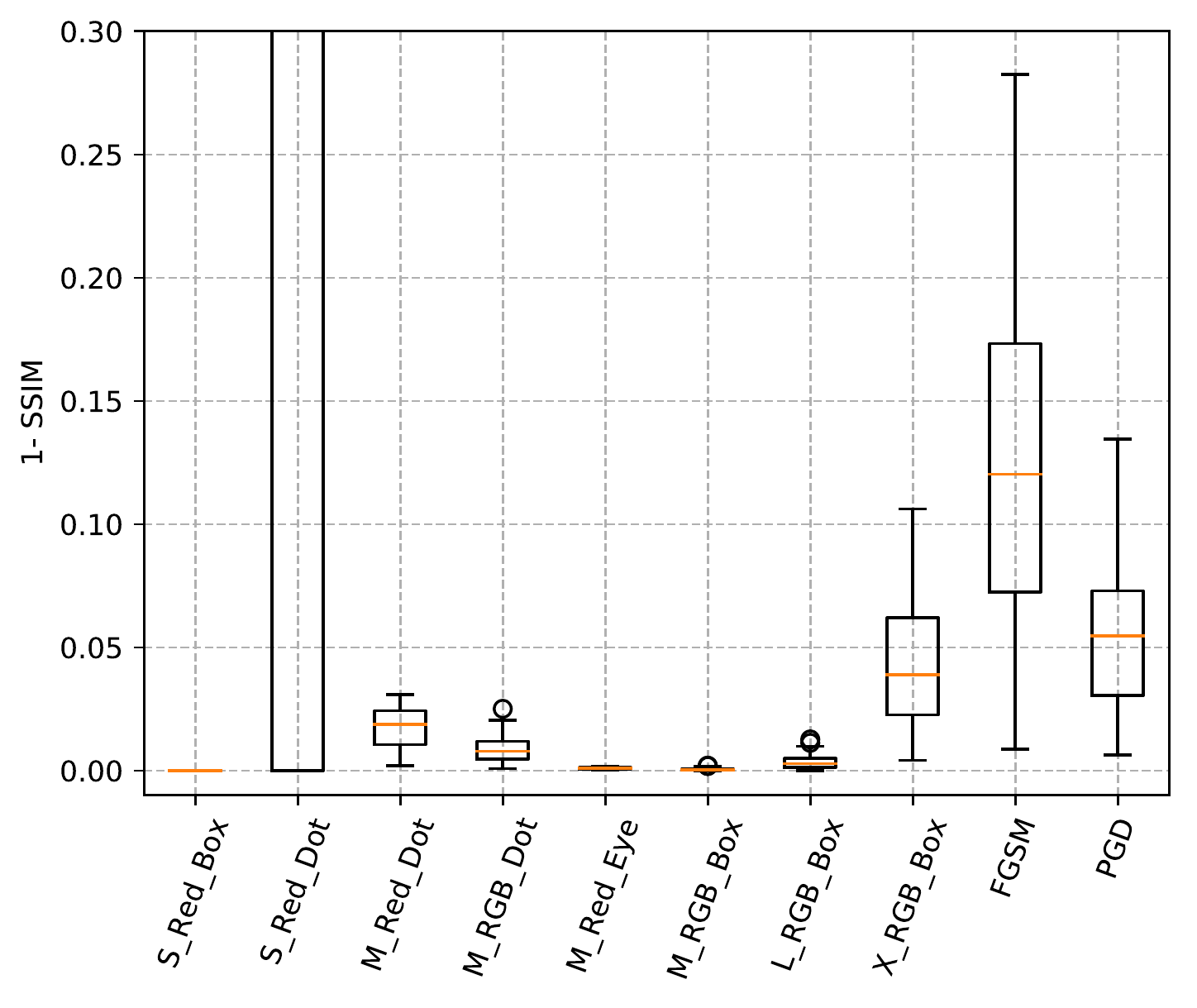} &
\includegraphics[width=0.26\textwidth]{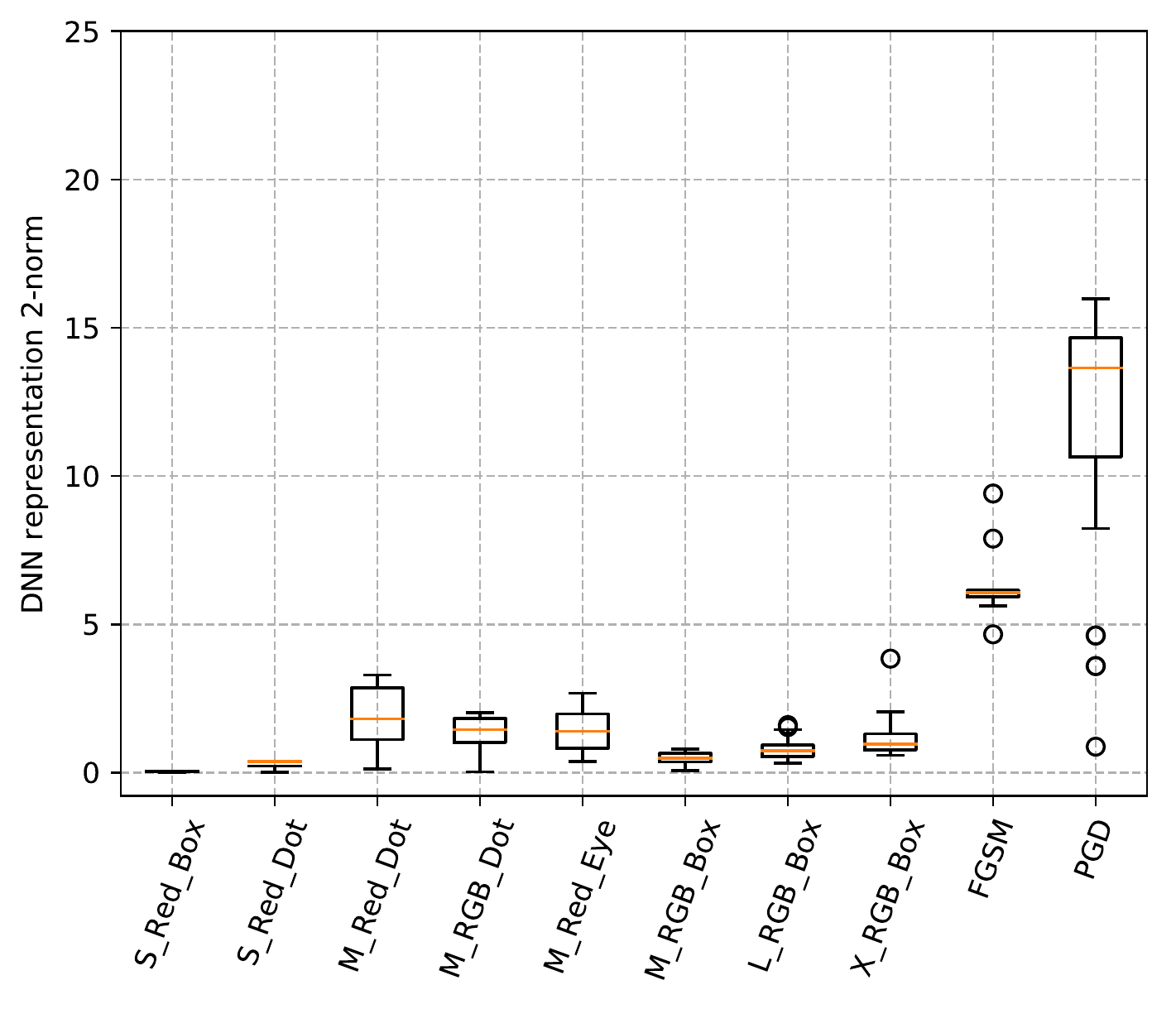} &
\includegraphics[width=0.26\textwidth]{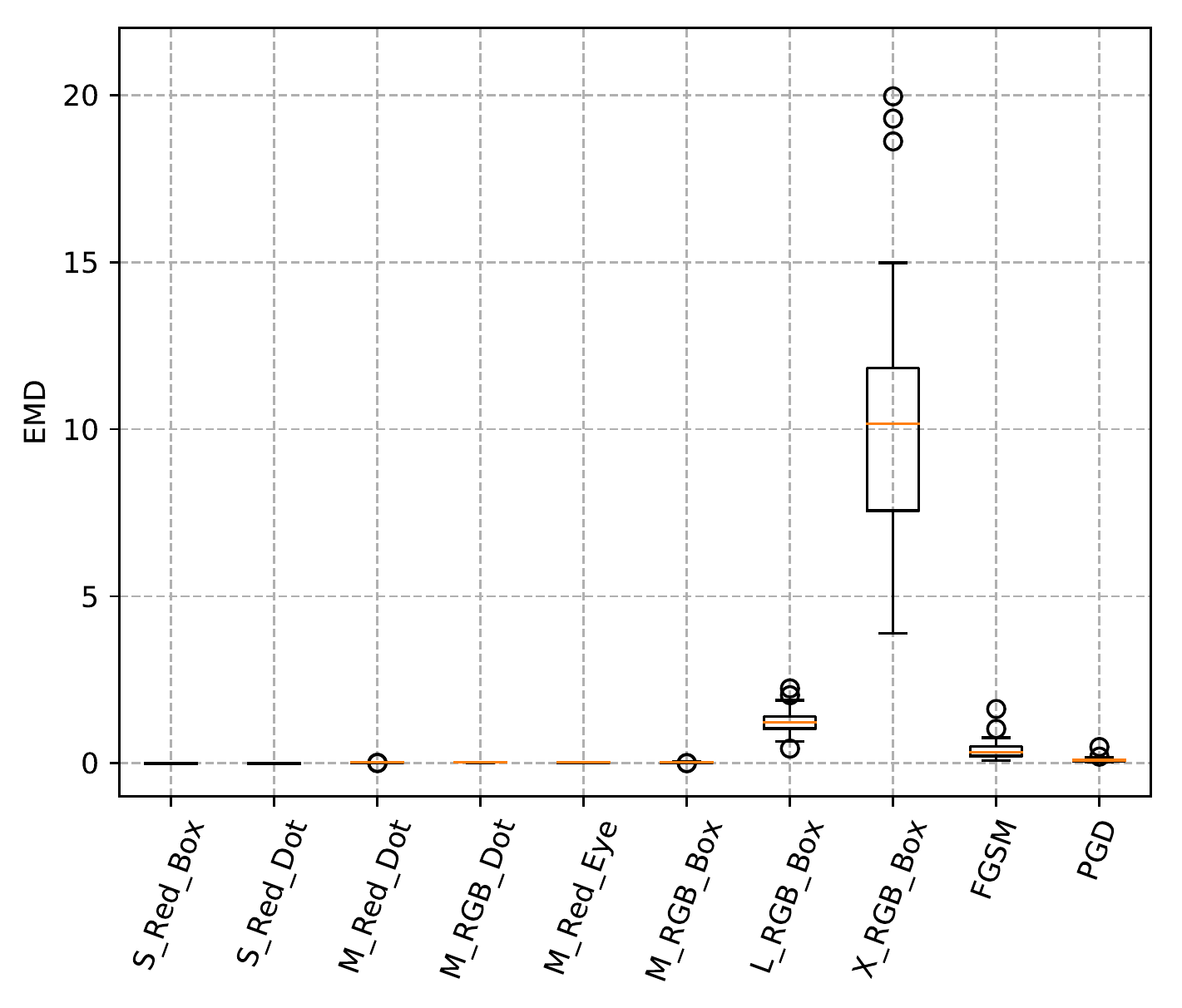}
\end{tabular}
\end{center}
\caption{Box plots of different measures $\rho$ on human JND images.}
\label{fig:measures}
\end{figure}

\textbf{Earth mover's distance} (EMD), also known as Wasserstein distance, is a distance function defined between two probability distributions on a given metric space.
The metric computes the minimum cost of converting one distribution to the other one.
EMD has been used as a distance metric in the image space also, e.g. for image retrieval~\cite{rubner2000earth}.
Given two images $\bfx_0$ and $\bfx$, EMD is calculated as
$\EMD(\bfx_0, \bfx) = \inf_{\gamma \in \Gamma(\bfx_0,\bfx)}\int_{\R \times \R} |a - b|d\gamma(a, b).$
Here, $\Gamma(\bfx_0, \bfx)$ is the set of joint distributions whose marginals are $\bfx_0$ and $\bfx$ (treated as histograms), respectively.
We test $\rho(\bfx, \bfx_0) \coloneqq \EMD(\bfx, \bfx_0)$ which assumes that the same amount of earth moving in images corresponds to the same detectability by human perception.
Figure~\ref{fig:measures}(right) shows the box plots of human JND images $\EMD(\bfx, \bfx_0)$ along different perturbation directions for $\bfx_0=\text{cat}$ (the full plots are in appendix Figure~\ref{fig:emd}).
It is immediately clear that on perturbation direction X-RGB-Box humans need to move a lot more earth as measured by EMD before they perceive the image difference.
These should not happen: ideally human JND should occur at the same $\rho(\bfx,\bfx_0)$ value.
The following test implies EMD probably should not be used to define adversarial attack detectability.
\begin{hypotest}
	$H_0$: Human JND images' $\EMD(\bfx, \bfx_0)$ on directions $\bfv_1=\text{M-RGB-Box}$, $\bfv_2=\text{X-RGB-Box}$ for $\bfx_0=\text{cat}$ have the same distribution.
	KS test yields p-value $6.4\times 10^{-19}$, rejecting $H_0$.
\end{hypotest}

\textbf{Structural Similarity} (SSIM) is intended to be a perceptual similarity measure that quantifies image quality loss due to compression~\cite{wang2004image},
and used as a signal fidelity measure with respect to humans in multiple research works~\cite{wang2009mean,sheikh2006statistical}.
SSIM has three elements: luminance, contrast and similarity of local structure.
Given two images $\bfx_0$ and $\bfx$, SSIM is defined by
$
	\SSIM(\bfx_0, \bfx) = \left(\frac{2\mu_{\bfx_0}\mu_{\bfx} + C_1}{\mu_{\bfx_0}^2 + \mu_{\bfx}^2 + C_1}\right)
	\left(\frac{2\sigma_{\bfx_0} \sigma_{\bfx} + C_2}{\sigma_{\bfx_0}^2 + \sigma_{\bfx}^2 + C_2}\right)
	\left(\frac{\sigma_{\bfx_0\bfx} + C_3}{\sigma_{\bfx_0}\sigma_{\bfx} + C_3}\right).
$
$\mu_{\bfx_0}$ and $\mu_{\bfx}$ are the sample means;
$\sigma_{\bfx_0}$, $\sigma_{\bfx}$ and $\sigma_{\bfx_0\bfx}$ are the standard deviation and sample cross correlation of $\bfx_0$ and $\bfx$ (after subtracting the mean) respectively.
To compute SSIM we use window size $7$ without Gaussian weights. 
Since SSIM is a similarity score, we define $\rho(\bfx, \bfx_0) = 1 - \SSIM(\bfx, \bfx_0)$. 
Figure~\ref{fig:measures}(left) shows the box plots of $1 - \SSIM(\bfx, \bfx_0)$ of our participant data for $\bfx_0=\text{panda}$ (The full plot is in appendix Figure~\ref{fig:SSIM}).
The following test implies 1 - SSIM probably should not be used to define adversarial attack detectability.
\begin{hypotest}
	$H_0$: Human JND $1-\SSIM(\bfx, \bfx_0)$ on directions $\bfv_1=\text{X-RGB-Box}$, $\bfv_2=\text{FGSM}$ for $\bfx_0=\text{panda}$ have the same distribution.
	KS test yields p-value $1.1\times 10^{-9}$, rejecting $H_0$.
\end{hypotest}

\textbf{Deep neural network (DNN) representation}.
Even though DNNs are designed with engineering goals in mind, studies comparing their internal representations to primate brains have found similarities~\cite{kriegeskorte2015deep}.
Let $\xi(\bfx) \in \R^D$ denote the last hidden layer representation of input image $\bfx$ in a DNN.
We may define $\rho(\bfx, \bfx_0) = \|\xi(\bfx) - \xi(\bfx_0)\|_p$ as a potential distance metric for our purpose.
We use Inception V3 representations with $D=2048$. 
Figure~\ref{fig:measures}(center) shows the box plots of human JND images' DNN 2-norm along different perturbation directions for $\bfx_0=\text{macaw}$.
The full plot for DNN $p=1,2,\infty$ norms and all animals is in appendix Figure~\ref{fig:embedding_Lp}.
Interestingly, the human JND images along the adversarial perturbation directions (FGSM and PGD) have much larger DNN $p$-norm than the $\pm 1$ perturbation directions.
As an example, $\bfx_0=\text{macaw}$, $\bfv_1=\text{M-Red-Dot}$ human JND images have median DNN 2-norm 1.8, while $\bfv_2=\text{PGD}$ human JND images have median 13.6.  
The following test implies that $2$-norm on DNN representation probably should not be used to define adversarial attack detectability.
\begin{hypotest}
	$H_0$: Human JND images' DNN 2-norm along $\bfv_1=\text{M-Red-Dot}$ and $\bfv_2=\text{PGD}$ for $\bfx_0=\text{macaw}$ have the same distribution.
	KS test yields p-value $2.1\times 10^{-12}$, rejecting $H_0$.
\end{hypotest}

\section{But which measure is a better approximation?}
\label{sec:approx}

We emphasize that our human experiments do \emph{not} support pixel $p$-norm, EMD, 1 - SSIM, or DNN representation as the \emph{correct} measure $\rho$.
Nonetheless, some of them may be \emph{useful} as computational {approximations} to human perception. 
As such, does our data suggest which measure offers the best approximation?
While none of the measures exactly satisfies $\forall \bfx_1, \bfx_2 \in J(\bfx_0), \rho(\bfx_1, \bfx_0)=\rho(\bfx_2,\bfx_0)$, the equality inspires the following idea:
the best measure should minimize the standard deviation of $\rho(\bfx, \bfx_0)$ over all human JND images $\bfx \in J(\bfx_0)$.
This is because $\rho(\bfx, \bfx_0)$ would have been a constant if the equality were true.
However, different measures have vastly different scales (e.g. for $p$-norms alone, the all-1 vector in $R^d$ has 1-norm $d$, 2-norm $\sqrt{d}$, and $\infty$-norm 1), making a direct comparison difficult.
Instead, we normalize by the center image $\bfx_0$ in order to find the best approximation: 
$\min_{\rho} \mathrm{std}\left({ \rho(\bfx, \bfx_0)  \over \rho(\bfx_0, \bm 0) }\right)$
where $\bm 0$ is the zero vector.
The standard deviation is taken over all our human experiment data for a particular center image $\bfx_0$, pooling all participants and all perturbation directions together, excluding ``give ups''.
Figure~\ref{fig:var} shows $\mathrm{std}\left({ \rho(\bfx, \bfx_0)  \over \rho(\bfx_0, \bm 0) }\right)$ of different measures.   For pixel $p$-norms this is presented as a function of $p$; EMD and 1 - SSIM are constant lines; and DNN has values larger than 0.12 for all $p$ and thus not shown (see appendix Figure~\ref{fig:varDNN} with DNN).

\begin{figure}[h]
	\begin{center}
		\begin{tabular}{ccc}
			\includegraphics[width=0.26\textwidth]{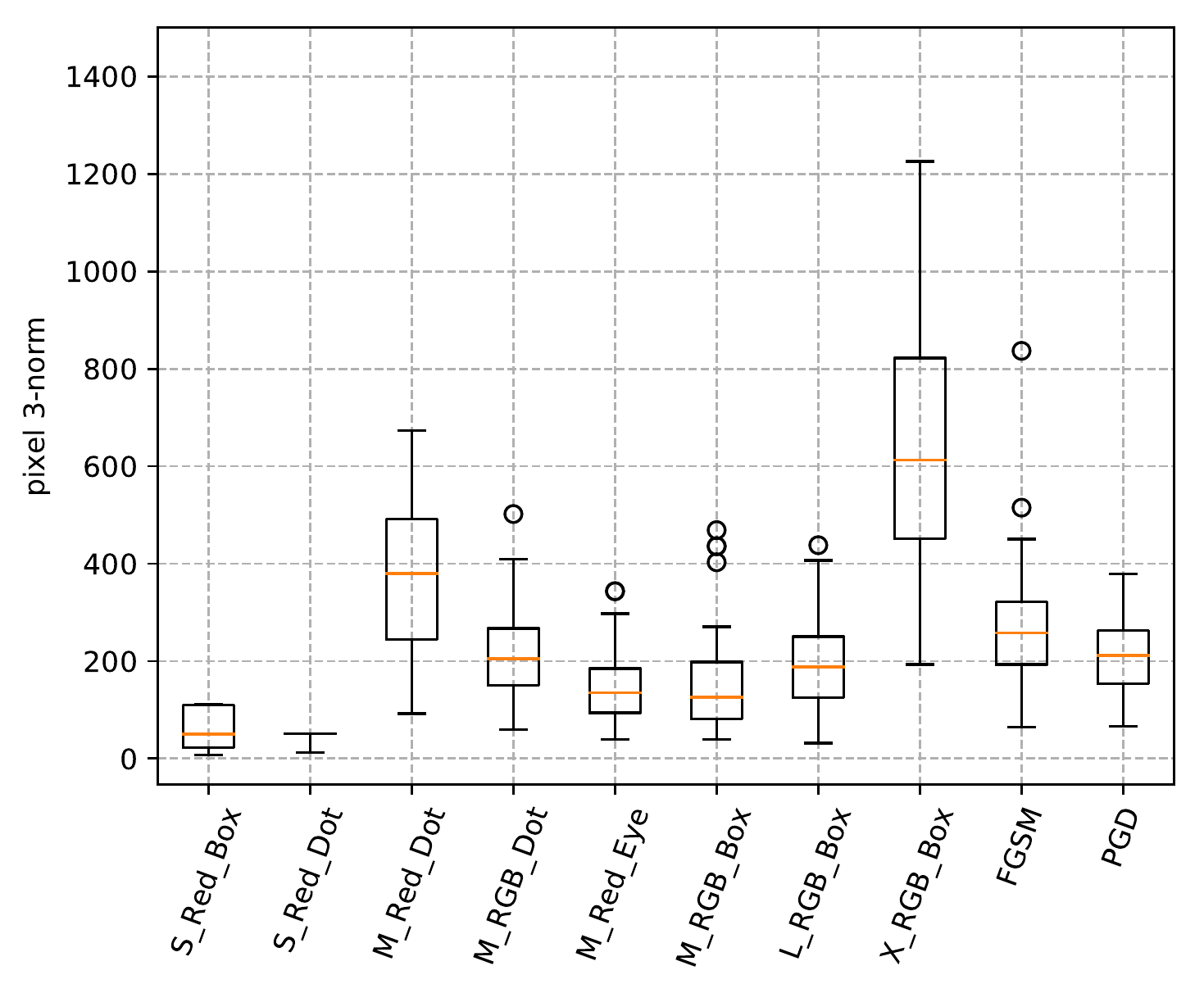} &
			\includegraphics[width=0.26\textwidth]{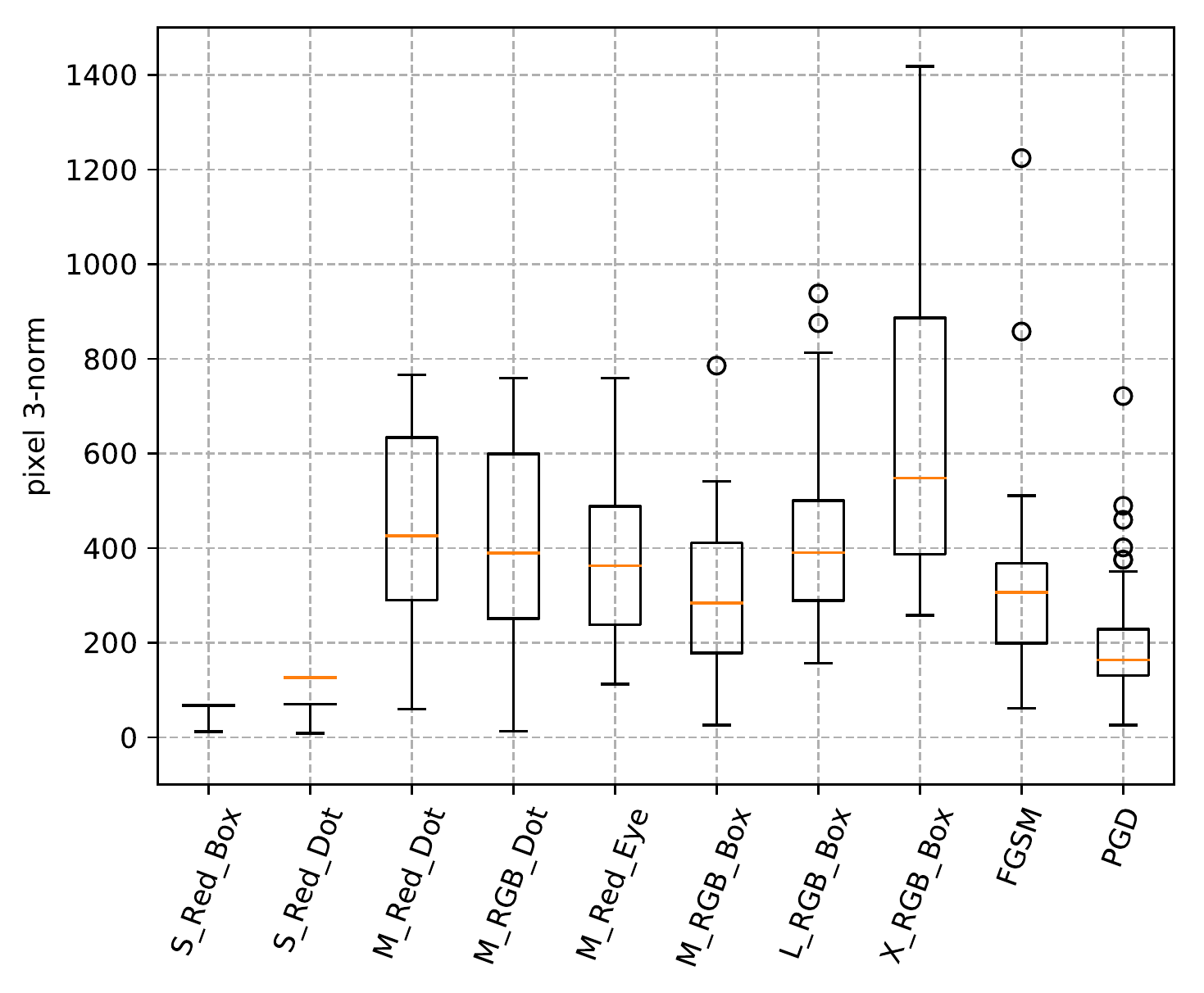} &
			\includegraphics[width=0.26\textwidth]{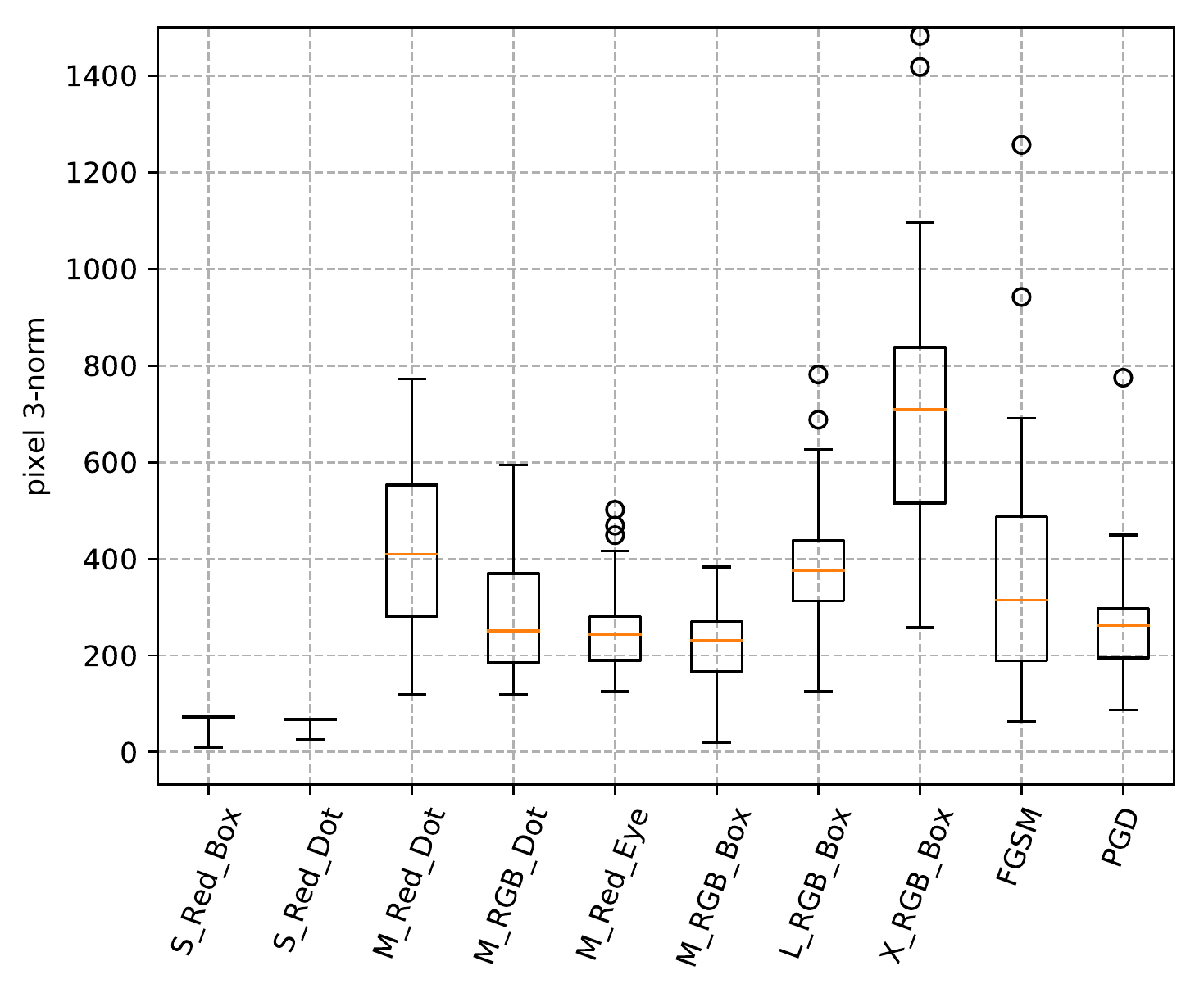} 
		\end{tabular}
	\end{center}
	\caption{
		Participants JND pixel 3-norm $\|\bfx-\bfx_0\|_3$ for panda, macaw, and cat, respectively.} 
	\label{fig:L3}
\end{figure}

Interestingly, by this criterion the pixel 3-norm is the best approximation of human JND judgment among the tested measures.
We plot $\|\bfx-\bfx_0\|_3$ of the human JND $\bfx$'s in Figure~\ref{fig:L3}.
Compared to pixel 1, 2, and $\infty$ norms in Figure~\ref{fig:Lp_excerpt}, 
EMD, 1 - SSIM, and DNN $p$-norm in Figure~\ref{fig:measures}, and their full plots in the appendix,
the median of pixel 3-norm (orange lines) are closer to having the same height. 
This qualitatively supports pixel 3-norm as a better approximation than the other measures.

%
%
%

\section{Conclusion}
Our behavioral experiment suggests that pixel $p$-norms, EMD, 1 - SSIM, and DNN representation $p$-norms do not match how humans judge just-noticeably-different images.
Even though pixel 3-norm is the closest approximation we tested, Figure~\ref{fig:L3} still contains significant variability.
Future research is needed to identify better measures of cognitive response to image distortion, and to generalize our work to other domains such as audio and text.

\subsubsection*{Acknowledgments}
The authors would like to thank Po-Ling Loh and Tim Rogers for helpful discussions.  This work is supported in part by NSF 1545481, 1561512, 1623605, 1704117, 1836978, the MADLab AF Center of Excellence FA9550-18-1-0166, American Family Insurance, and the University of Wisconsin.

\bibliography{ms}

\begin{thebibliography}{10}

\bibitem{athalye2018obfuscated}
Anish Athalye, Nicholas Carlini, and David Wagner.
\newblock Obfuscated gradients give a false sense of security: Circumventing
  defenses to adversarial examples.
\newblock {\em arXiv preprint arXiv:1802.00420}, 2018.

\bibitem{buhrmester2011amazon}
Michael Buhrmester, Tracy Kwang, and Samuel~D Gosling.
\newblock Amazon's mechanical turk: A new source of inexpensive, yet
  high-quality, data?
\newblock {\em Perspectives on Psychological Science}, 6(1):3--5, 2011.

\bibitem{carlini2017adversarial}
Nicholas Carlini and David Wagner.
\newblock Adversarial examples are not easily detected: Bypassing ten detection
  methods.
\newblock In {\em Proceedings of the 10th ACM Workshop on Artificial
  Intelligence and Security}, pages 3--14. ACM, 2017.

\bibitem{deng2009imagenet}
Jia Deng, Wei Dong, Richard Socher, Li-Jia Li, Kai Li, and Li~Fei-Fei.
\newblock Imagenet: A large-scale hierarchical image database.
\newblock In {\em Proceedings of the IEEE Conference on Computer Vision and
  Pattern Recognition}, pages 248--255. Ieee, 2009.

\bibitem{fechner1966elements}
Gustav~Theodor Fechner, Edwin~Garrigues Boring, Davis~H Howes, and Helmut~E
  Adler.
\newblock {\em Elements of Psychophysics. Translated by Helmut E. Adler. Edited
  by Davis H. Howes And Edwin G. Boring, With an Introd. by Edwin G. Boring}.
\newblock Holt, Rinehart and Winston, 1966.

\bibitem{goodfellow6572explaining}
Ian~J Goodfellow, Jonathon Shlens, and Christian Szegedy.
\newblock Explaining and harnessing adversarial examples (2014).
\newblock {\em arXiv preprint arXiv:1412.6572}, 2014.

\bibitem{itti2001computational}
Laurent Itti and Christof Koch.
\newblock Computational modelling of visual attention.
\newblock {\em Nature Reviews: Neuroscience}, 2(3):194, 2001.

\bibitem{jamieson2015next}
Kevin~G Jamieson, Lalit Jain, Chris Fernandez, Nicholas~J. Glattard, and Rob
  Nowak.
\newblock Next: A system for real-world development, evaluation, and
  application of active learning.
\newblock In C.~Cortes, N.~D. Lawrence, D.~D. Lee, M.~Sugiyama, and R.~Garnett,
  editors, {\em Advances in Neural Information Processing Systems 28}, pages
  2656--2664. Curran Associates, Inc., 2015.

\bibitem{kriegeskorte2015deep}
Nikolaus Kriegeskorte.
\newblock Deep neural networks: a new framework for modeling biological vision
  and brain information processing.
\newblock {\em Annual Review of Vision Science}, 1:417--446, 2015.

\bibitem{krizhevsky2009learning}
Alex Krizhevsky and Geoffrey Hinton.
\newblock Learning multiple layers of features from tiny images.
\newblock Technical report, Citeseer, 2009.

\bibitem{lecun1998mnist}
Yann LeCun.
\newblock The mnist database of handwritten digits.
\newblock \url{http://yann.lecun.com/exdb/mnist/}, 1998.

\bibitem{madry2017towards}
Aleksander Madry, Aleksandar Makelov, Ludwig Schmidt, Dimitris Tsipras, and
  Adrian Vladu.
\newblock Towards deep learning models resistant to adversarial attacks.
\newblock {\em arXiv preprint arXiv:1706.06083}, 2017.

\bibitem{moosavi2017universal}
Seyed-Mohsen Moosavi-Dezfooli, Alhussein Fawzi, Omar Fawzi, and Pascal
  Frossard.
\newblock Universal adversarial perturbations.
\newblock In {\em Proceedings of the IEEE Conference on Computer Vision and
  Pattern Recognition}, pages 1765--1773, 2017.

\bibitem{moosavi2016deepfool}
Seyed-Mohsen Moosavi-Dezfooli, Alhussein Fawzi, and Pascal Frossard.
\newblock Deepfool: a simple and accurate method to fool deep neural networks.
\newblock In {\em Proceedings of the IEEE Conference on Computer Vision and
  Pattern Recognition}, pages 2574--2582, 2016.

\bibitem{rensink2002change}
Ronald~A Rensink.
\newblock Change detection.
\newblock {\em Annual Review of Psychology}, 53(1):245--277, 2002.

\bibitem{rubner2000earth}
Yossi Rubner, Carlo Tomasi, and Leonidas~J Guibas.
\newblock The earth mover's distance as a metric for image retrieval.
\newblock {\em International Journal of Computer Vision}, 40(2):99--121, 2000.

\bibitem{salamati2019perception}
Mahmoud Salamati, Sadegh Soudjani, and Rupak Majumdar.
\newblock Perception-in-the-loop adversarial examples.
\newblock {\em arXiv preprint arXiv:1901.06834}, 2019.

\bibitem{Sharif2018Bright}
Mahmood Sharif, Lujo Bauer, and Michael~K. Reiter.
\newblock On the suitability of lp-norms for creating and preventing
  adversarial examples.
\newblock In {\em The Bright and Dark Sides of Computer Vision: Challenges and
  Opportunities for Privacy and Security (CVPR Workshop)}, 2018.

\bibitem{sheikh2006statistical}
Hamid~R Sheikh, Muhammad~F Sabir, and Alan~C Bovik.
\newblock A statistical evaluation of recent full reference image quality
  assessment algorithms.
\newblock {\em IEEE Transactions on Image Processing}, 15(11):3440--3451, 2006.

\bibitem{sievert2017next}
Scott Sievert, Daniel Ross, Lalit Jain, Kevin Jamieson, Rob Nowak, and Robert
  Mankoff.
\newblock Next: A system to easily connect crowdsourcing and adaptive data
  collection.
\newblock In {\em Proceedings of the 16th Python in Science Conference}, pages
  113--119, 2017.

\bibitem{simons2005change}
Daniel~J Simons and Michael~S Ambinder.
\newblock Change blindness: Theory and consequences.
\newblock {\em Current Directions in Psychological Science}, 14(1):44--48,
  2005.

\bibitem{szegedy2016rethinking}
Christian Szegedy, Vincent Vanhoucke, Sergey Ioffe, Jon Shlens, and Zbigniew
  Wojna.
\newblock Rethinking the inception architecture for computer vision.
\newblock In {\em Proceedings of the IEEE Conference on Computer Vision and
  Pattern Recognition}, pages 2818--2826, 2016.

\bibitem{szegedy2013intriguing}
Christian Szegedy, Wojciech Zaremba, Ilya Sutskever, Joan Bruna, Dumitru Erhan,
  Ian Goodfellow, and Rob Fergus.
\newblock Intriguing properties of neural networks.
\newblock {\em arXiv preprint arXiv:1312.6199}, 2013.

\bibitem{wang2009mean}
Zhou Wang and Alan~C Bovik.
\newblock Mean squared error: Love it or leave it? a new look at signal
  fidelity measures.
\newblock {\em IEEE Signal Processing Magazine}, 26(1):98--117, 2009.

\bibitem{wang2004image}
Zhou Wang, Alan~C Bovik, Hamid~R Sheikh, and Eero~P Simoncelli.
\newblock Image quality assessment: from error visibility to structural
  similarity.
\newblock {\em IEEE transactions on Image Processing}, 13(4):600--612, 2004.

\bibitem{wolfe2010visual}
Jeremy~M Wolfe.
\newblock Visual search.
\newblock {\em Current Biology}, 20(8):R346--R349, 2010.

\bibitem{zhang2008just}
Xiaohui Zhang, Weisi Lin, and Ping Xue.
\newblock Just-noticeable difference estimation with pixels in images.
\newblock {\em Journal of Visual Communication and Image Representation},
  19(1):30--41, 2008.

\end{thebibliography}
\bibliographystyle{plain}

\newpage
\appendix
\section{Supplemental materials}

\subsection{Further statistical tests}

We also take $\bfx_0=\text{panda}$ and look at the two perturbation directions $\bfv_1=\text{M-Red-Dot}$ and $\bfv_2=\text{M-RGB-Dot}$.
The directions again have the same sparsity; this time they also share the same ``shape of support'': the nonzero elements of $\bfv_1, \bfv_2$ are both randomly scattered over pixels.  The difference is that $\bfv_1$ changes only the red color channel on 288 random pixels, while $\bfv_2$ changes all three channels but only on $288/3=96$ random pixels.
Implication~\ref{imp:same_s} again predicts that the humans should reach just-noticeable-difference at the same scale.
But Figure~\ref{fig:boxplot_a}(left) suggest that humans are more sensitive to simultaneous changes to all RGB channels:
scales $a_1$, $a_2$ have median 57.5 and 31, respectively.

\begin{hypotest}
	The null hypothesis $H_0$ is: The sample $\{a_1^{(1)}, \ldots, a_1^{(n)}\}$ generated from $\bfx_0=\text{panda}$, $\bfv_1=\text{M-Red-Dot}$ and the sample $\{a_2^{(1)}, \ldots, a_2^{(n)}\}$ generated from $\bfx_0=\text{panda}$, $\bfv_2=\text{M-RGB-Dot}$ come from the same distribution.
	A two-sample Kolmogorov-Smirnov test on our data ($n=42$) yields a p-value $2.1\times 10^{-4}$, rejecting $H_0$.
\end{hypotest}

%


These two tests refute implication~\ref{imp:same_s}.
They already indicate that no $p^*$ can make the central hypothesis true.

While implication~\ref{imp:same_s} focuses on perturbation directions of the same sparsity, the next implication states that if one perturbation is changing more dimensions than the other, it should achieve just noticeable difference with a smaller perturbation scale.  Again, this is true for all $p$.
\begin{imp}
	\label{imp:diff_s}
	$\forall p>0$, $\forall \pm 1$-perturbation directions $\bfv_1, \bfv_2$ with sparsity $s_1 > s_2$,
	suppose $\exists a_1, a_2$ such that
	$\bfx_1 = \bfx_0 + a_1 \bfv_1 \in J(\bfx_0)$ and 
	$\bfx_2 = \bfx_0 + a_2 \bfv_2 \in J(\bfx_0)$. 
	Then $a_1 < a_2$.
\end{imp}

To further strengthen our case, we also test implication~\ref{imp:diff_s} which states that it is easier to notice changes if the perturbation $\bfv$ has larger support $s$.
This is mostly true as seen in Figure~\ref{fig:boxplot_a}: the median of $a$ generally decreases as $\bfv$ support size increases in the order of S, M, L, X.
However, there is a curious inversion on $\bfx_0=\text{panda}$, $\bfv_1=\text{L-RGB-Box}$ vs $\bfv_2=\text{X-RGB-Box}$:
Implication~\ref{imp:diff_s} predicts that $a_1 > a_2$, but human behaviors have mean 6.5 and 9.9 (and median 6 and 9.5), respectively: the other way around.
The human data for these perturbations are not censored; Figure~\ref{fig:boxplot_a} also suggests they are close to normal in distribution.
We therefore perform a one-tailed two-sample $t$-test with unequal variances. 

\begin{hypotest}
	The null hypothesis $H_0$ is: Human JND scales generated from $\bfx_0=\text{panda}$, $\bfv_1=\text{L-RGB-Box}$ has equal mean as those generated from $\bfx_0=\text{panda}$, $\bfv_2=\text{X-RGB-Box}$.
	The left-tailed alternative hypothesis $H_a$ is: the former has a smaller mean.
	A one-tailed two-sample $t$-test with unequal variances on our data $(n=42)$ yields a p-value of $1.8 \times 10^{-5}$, rejecting $H_0$ and retaining $H_1$. 
\end{hypotest}

The test suggests that the inversion is statistically significant, thus refuting implication~\ref{imp:diff_s}.
We speculate that the inversion is due to the black-and-white panda making the L box boundary more prominent, see Figure~\ref{fig:perturbed_images}(g). 

\subsection{Additional figures}

\begin{figure}[!htb]
	\centering
	\begin{subfigure}[t]{0.3\textwidth}
		\centering
		\includegraphics[width=\textwidth]{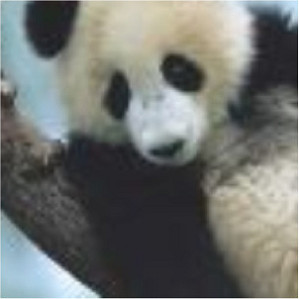}
		\caption{panda}
	\end{subfigure}
	\begin{subfigure}[t]{0.3\textwidth}
		\centering
		\includegraphics[width=\textwidth]{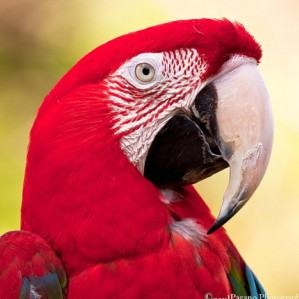}
		\caption{macaw}
	\end{subfigure}
	\begin{subfigure}[t]{0.3\textwidth}
		\centering
		\includegraphics[width=\textwidth]{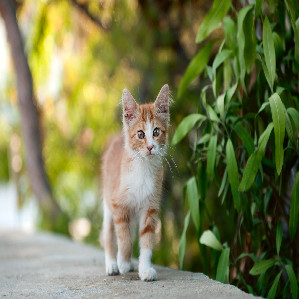}
		\caption{cat}
	\end{subfigure}
	\caption{The three natural images $\bfx_0$}~\label{fig:centers}
\end{figure}

\begin{figure}[htb]
	\centering
	\includegraphics[width=0.118\textwidth]{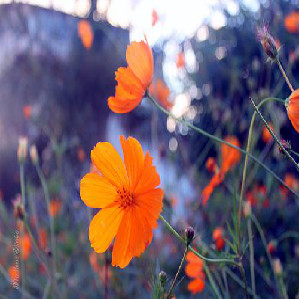}
	\includegraphics[width=0.118\textwidth]{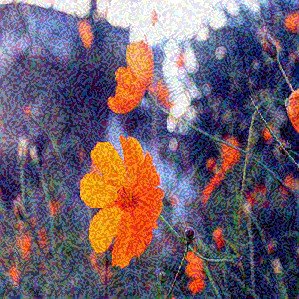}
	\includegraphics[width=0.118\textwidth]{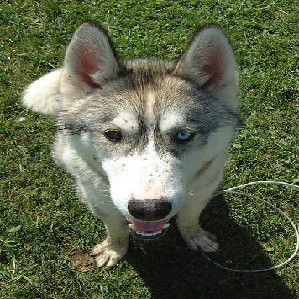}
	\includegraphics[width=0.118\textwidth]{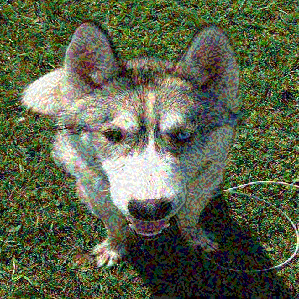}
	\includegraphics[width=0.118\textwidth]{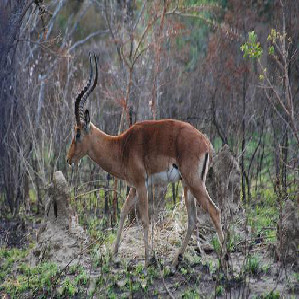}
	\includegraphics[width=0.118\textwidth]{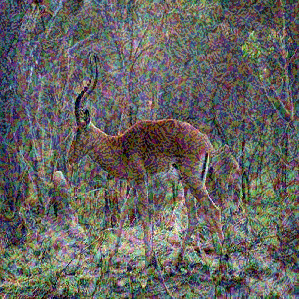}
	\includegraphics[width=0.118\textwidth]{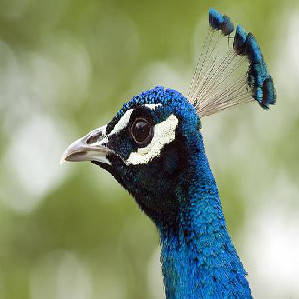}
	\includegraphics[width=0.118\textwidth]{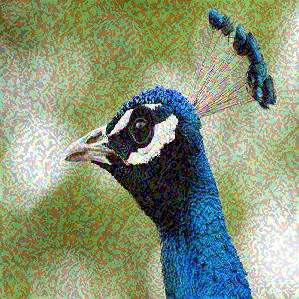}
	\caption{Photos used for guard trials along with their distorted version}~\label{fig:guard_rays_distorted}
\end{figure}

\begin{figure}[h]
	\begin{center}
		\begin{tabular}{ccc}
			panda & macaw & cat \\
			\includegraphics[width=0.25\textwidth]{figures/pixel_1_norm_boxplot_panda-eps-converted-to.pdf} &
			\includegraphics[width=0.25\textwidth]{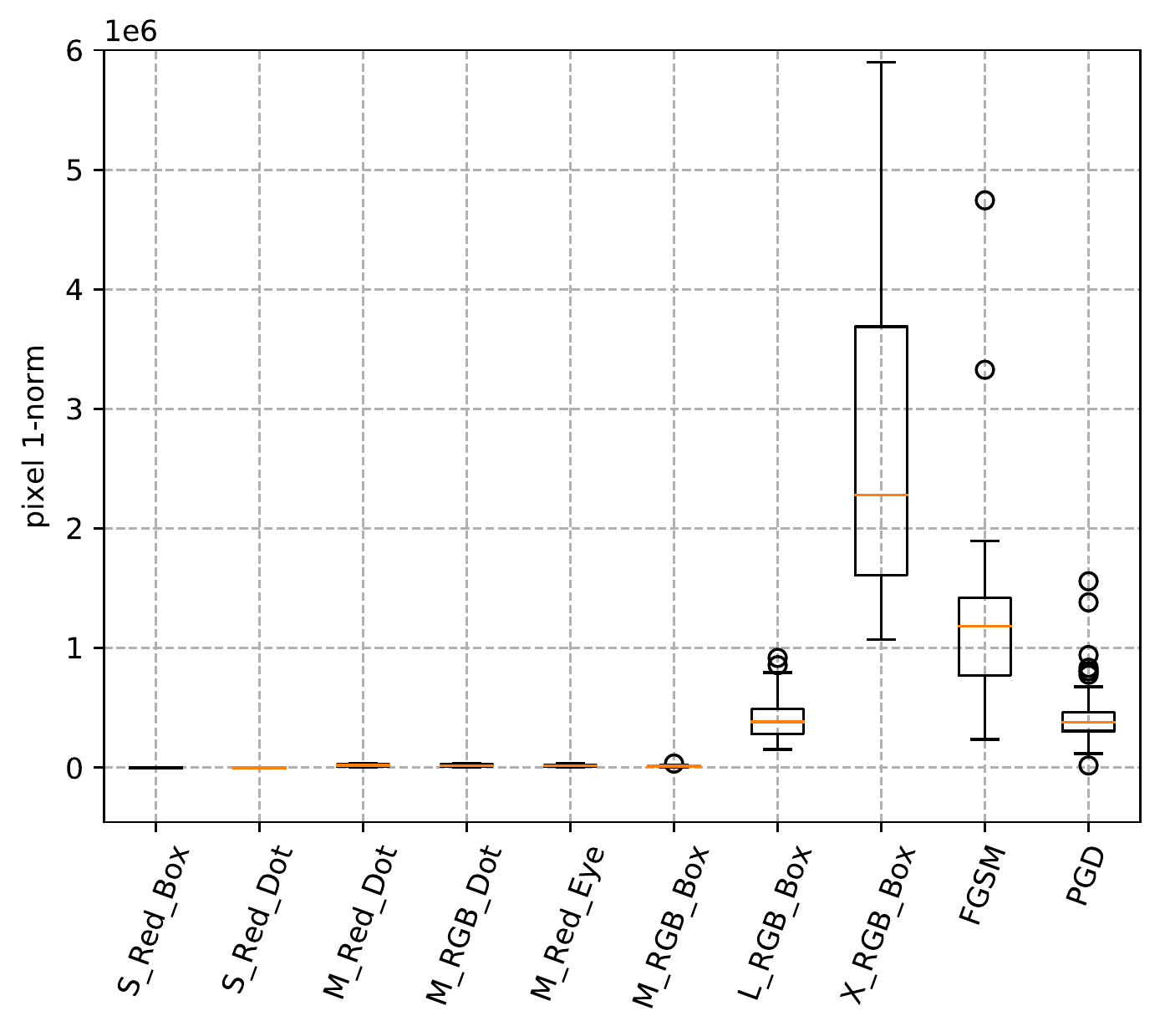} &
			\includegraphics[width=0.25\textwidth]{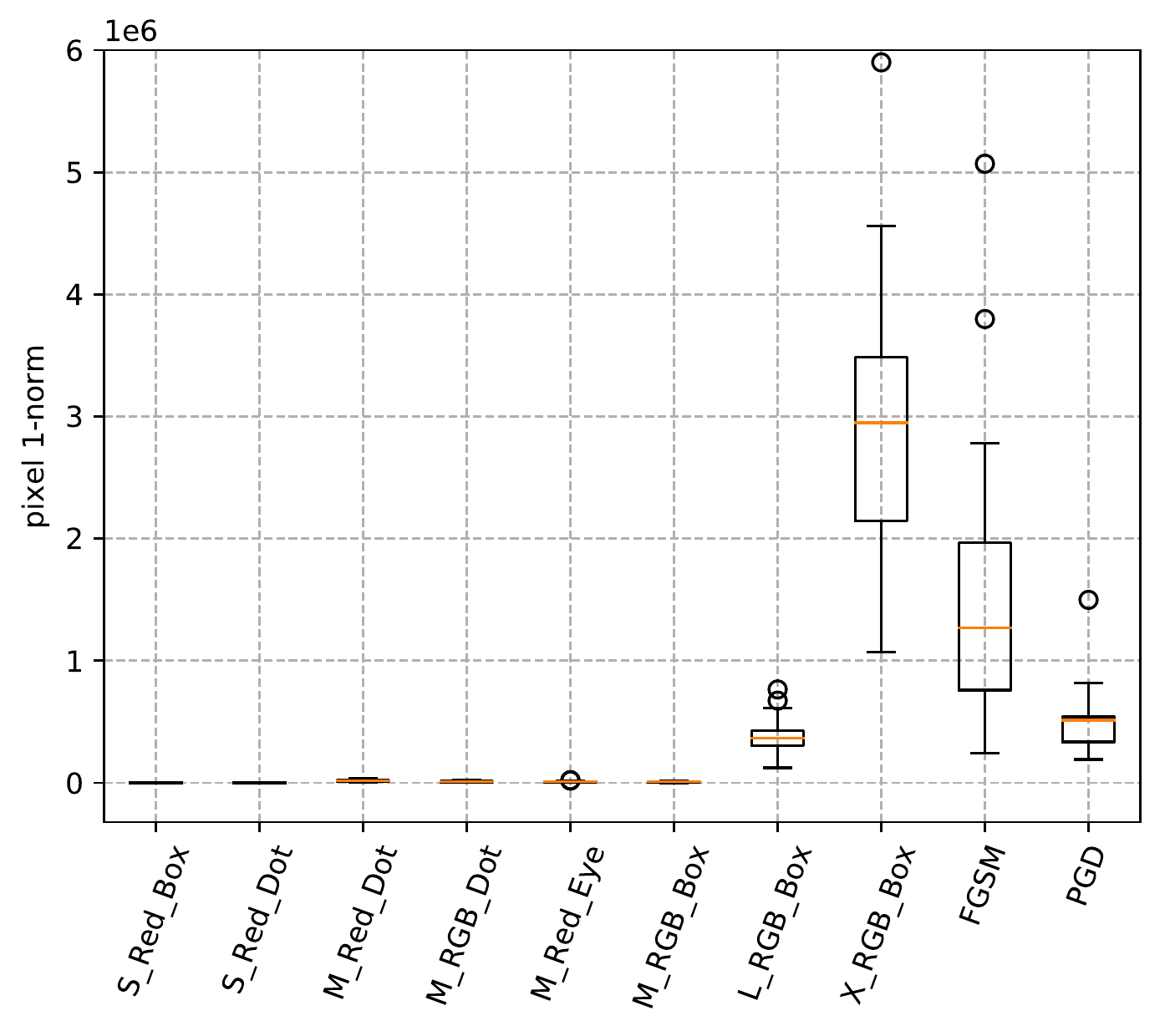} \\
			\includegraphics[width=0.25\textwidth]{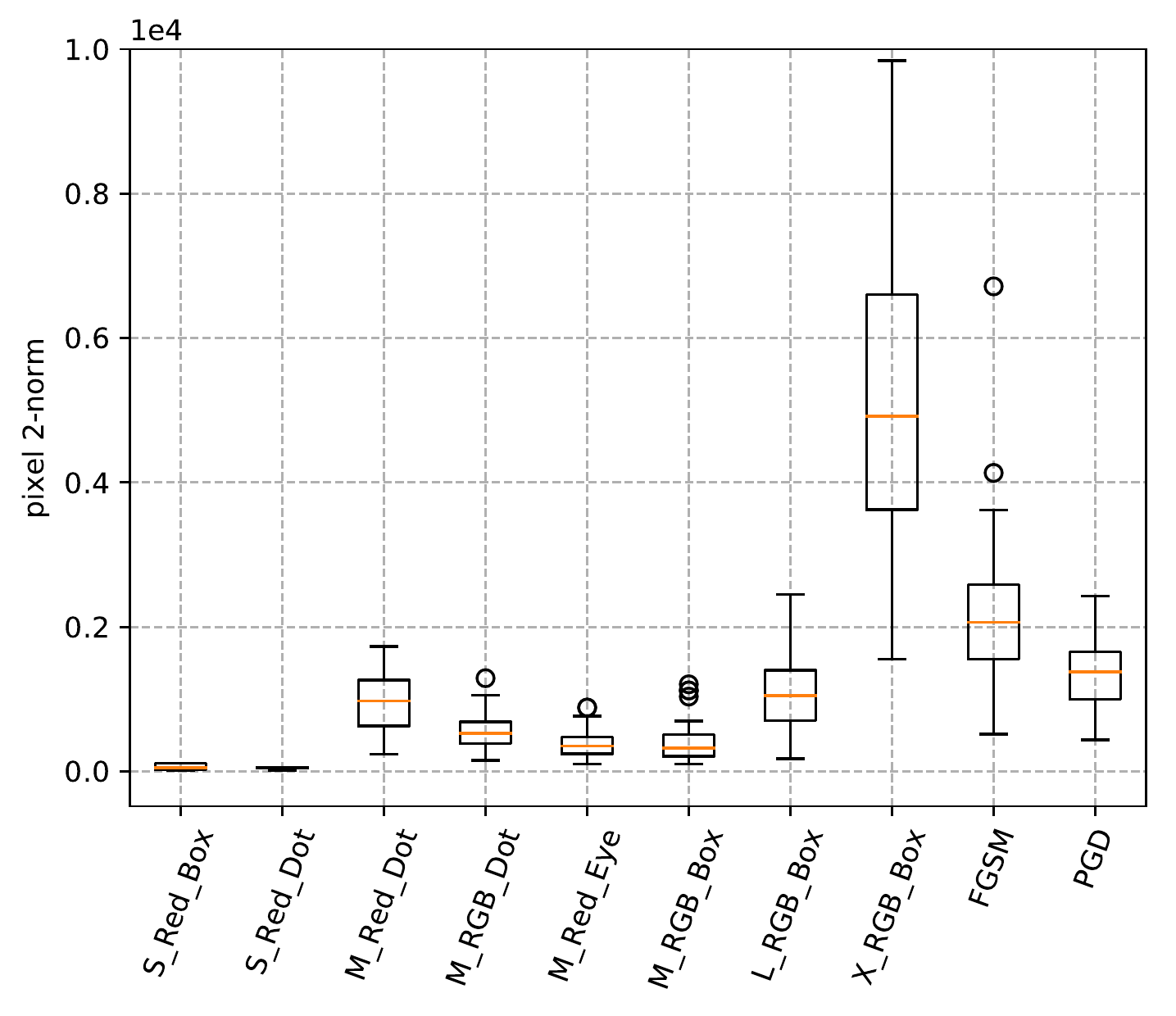} &
			\includegraphics[width=0.25\textwidth]{figures/pixel_2_norm_boxplot_macaw-eps-converted-to.pdf} &
			\includegraphics[width=0.25\textwidth]{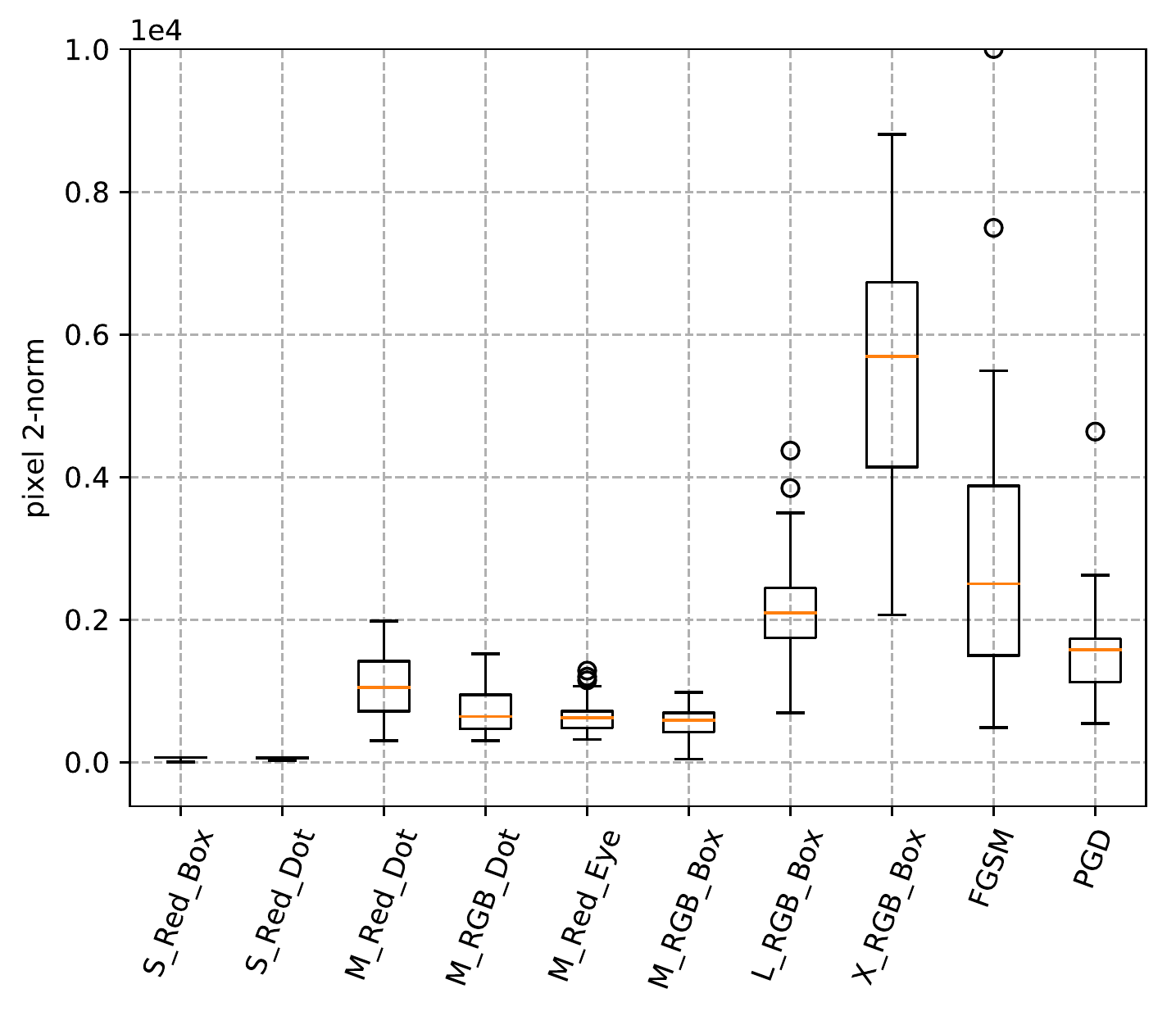} \\
			\includegraphics[width=0.25\textwidth]{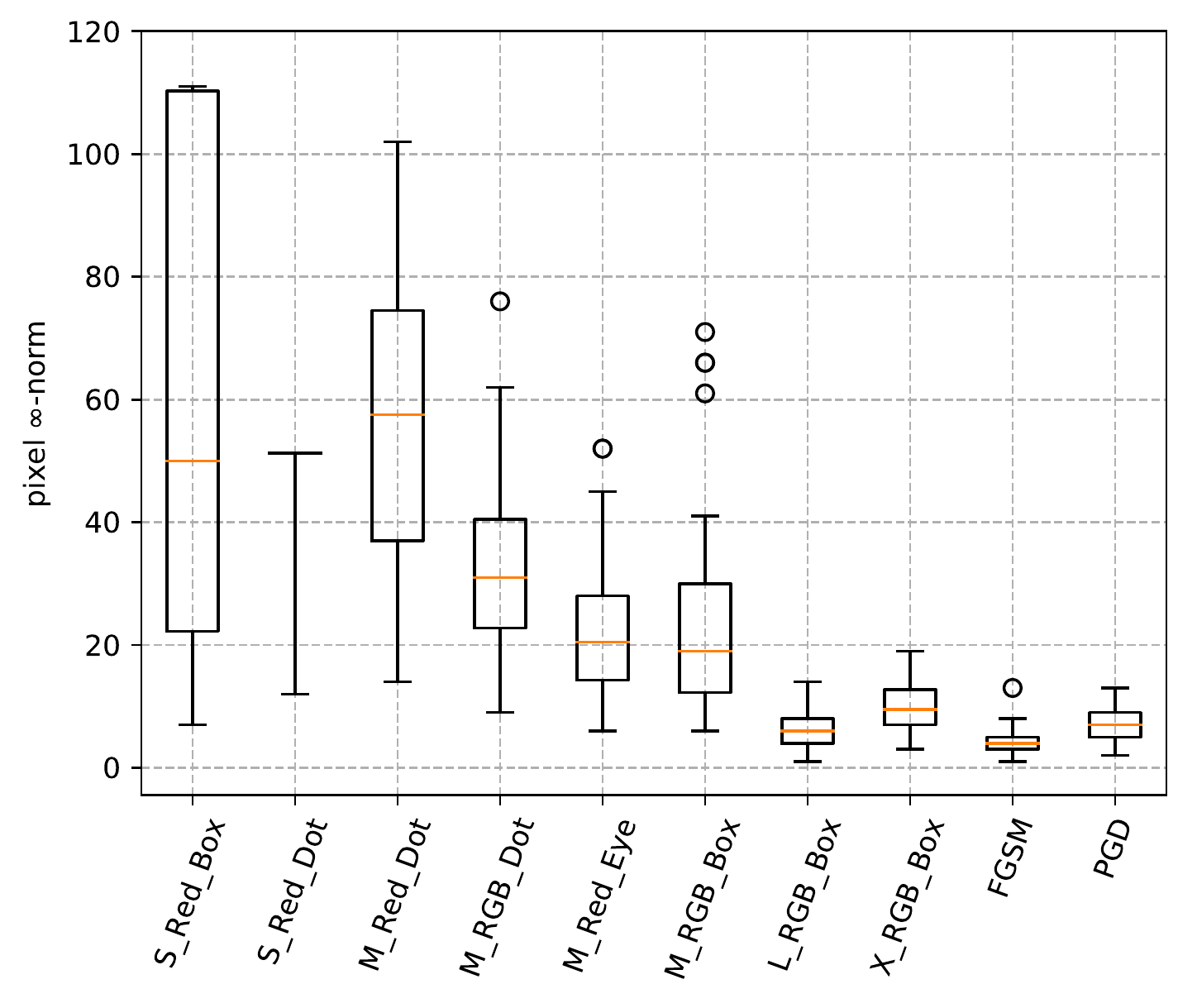} &
			\includegraphics[width=0.25\textwidth]{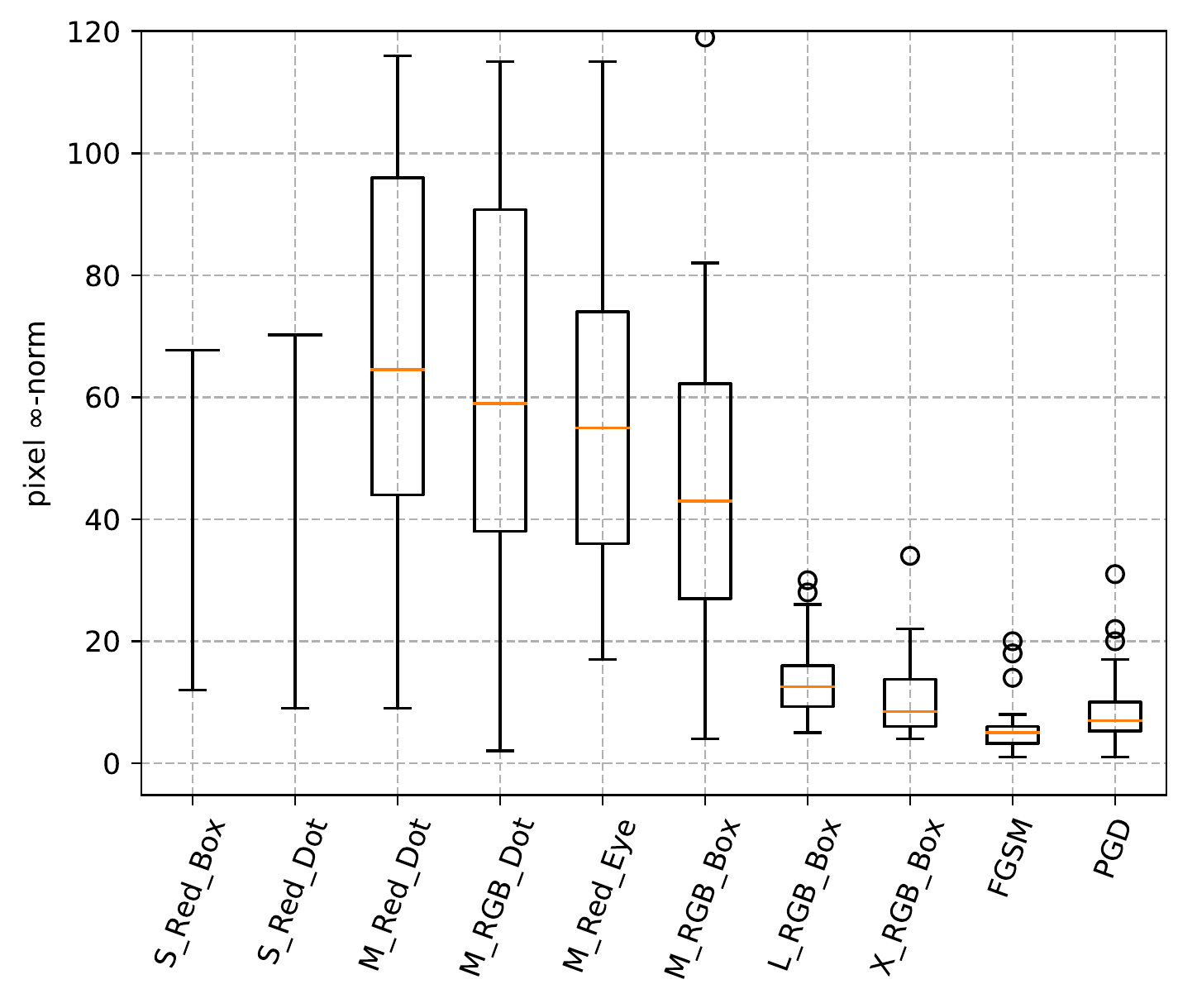} &
			\includegraphics[width=0.25\textwidth]{figures/pixel_inf_norm_boxplot_cat-eps-converted-to.pdf} 
		\end{tabular}
	\end{center}
	\caption{Participant JND $\bfx$'s pixel norm $\|\bfx-\bfx_0\|_p$ for $p=1$ (top row), 2 (middle row), $\infty$ (bottom row).
Within a plot, each vertical box is for a perturbation direction $\bfv$.
The box plot depicts the median, quartiles, and outliers. 
		If the central hypothesis were true, one expects a plot to have similar medians (orange lines).}
	\label{fig:Lp}
\end{figure}

\begin{figure}[!h]
	\centering
	\begin{subfigure}[t]{0.3\textwidth}
		\centering
		\includegraphics[width=\textwidth]{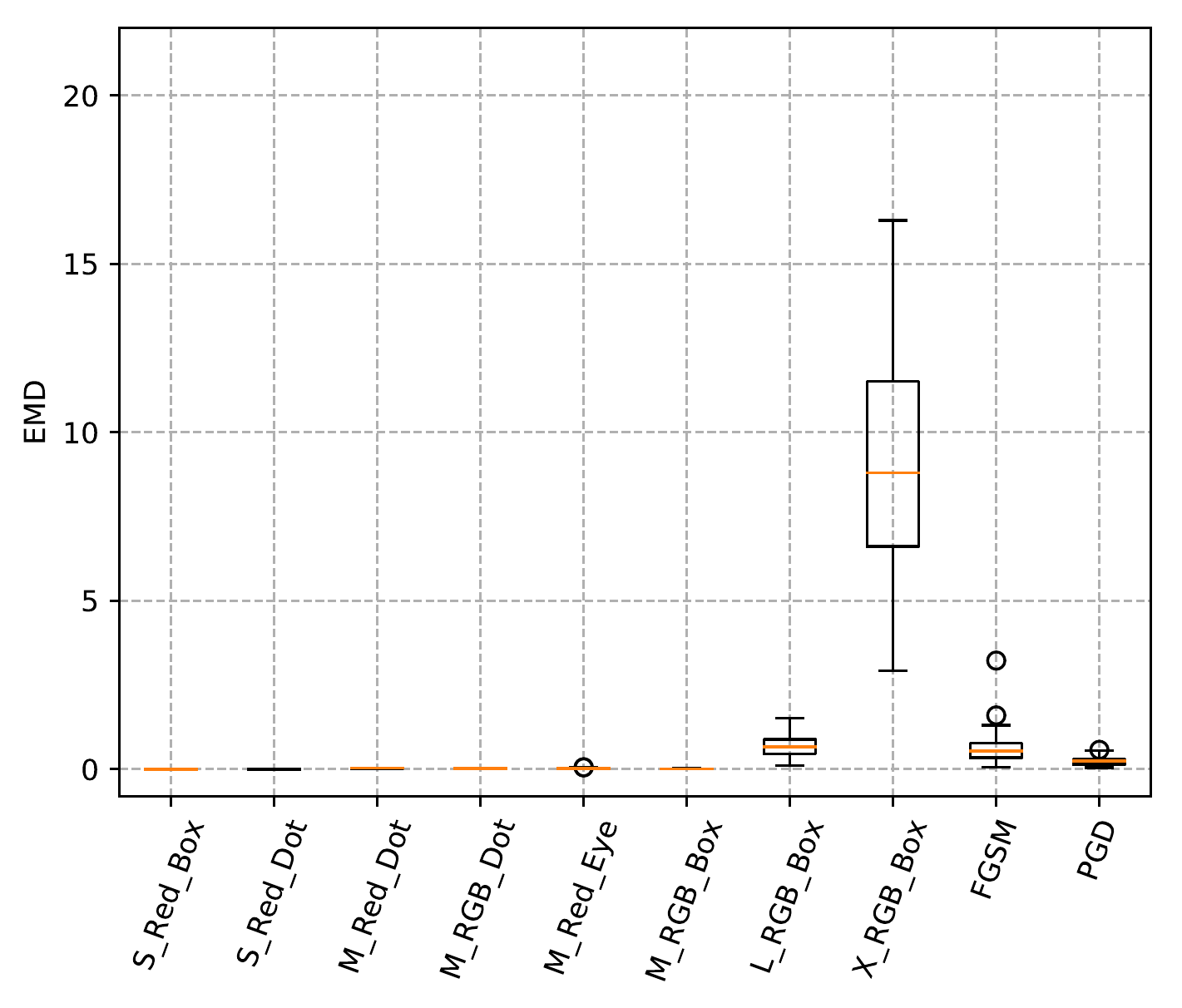}
		\caption{panda}
	\end{subfigure}
	\begin{subfigure}[t]{0.3\textwidth}
		\centering
		\includegraphics[width=\textwidth]{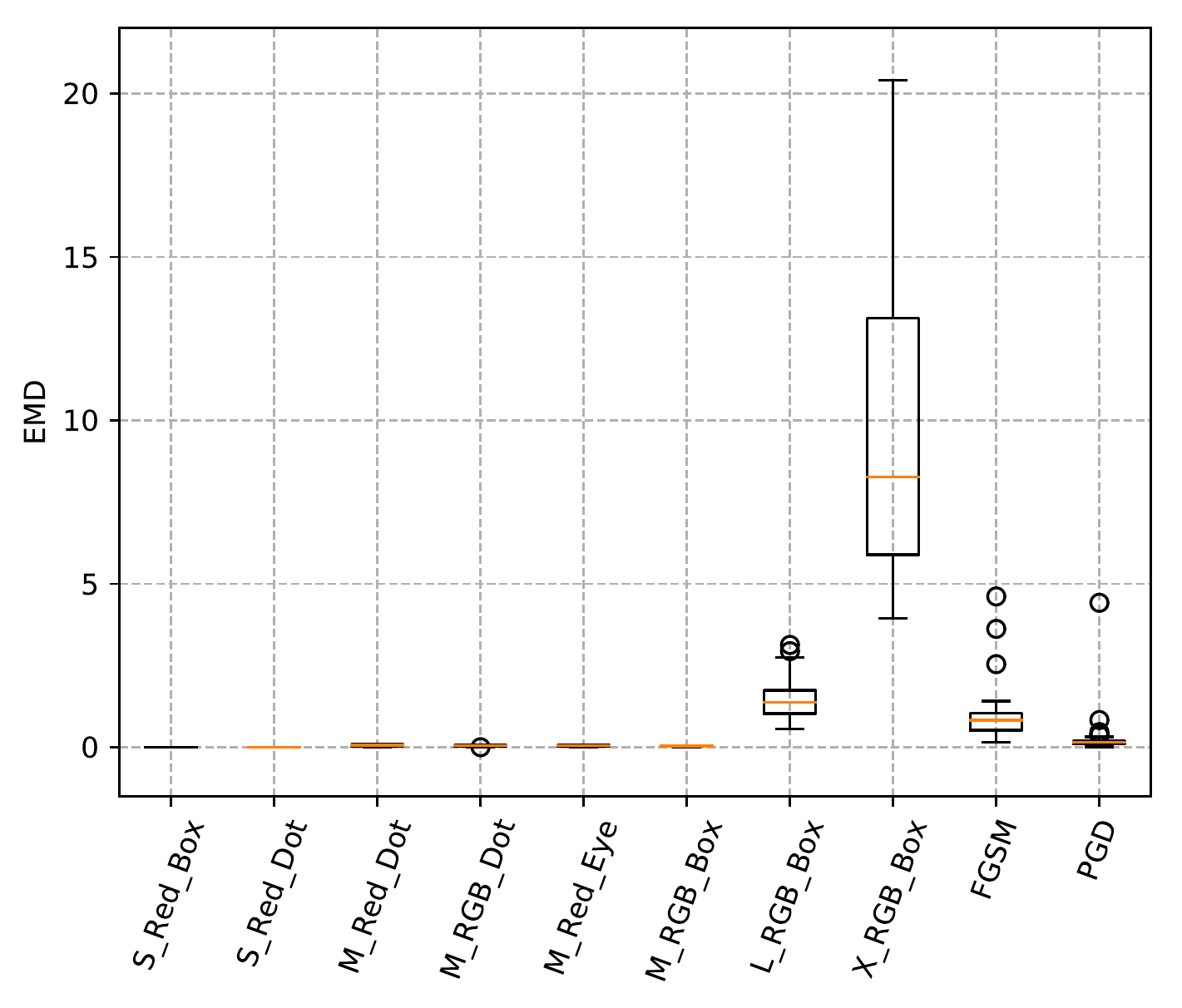}
		\caption{macaw}
	\end{subfigure}
	\begin{subfigure}[t]{0.3\textwidth}
		\centering
		\includegraphics[width=\textwidth]{figures/emd_boxplot_cat-eps-converted-to.pdf}
		\caption{cat}
	\end{subfigure}
	\caption{Box plots of Earth Mover's Distance on human JND images.
Recall for each natural image $\bfx_0$ and each perturbation direction $\bfv$, our $n$ participants decided which image $\bfx^{(j)}=\Pi(\bfx_0+a^{(j)} \bfv)$ is JND to them, for $j=1\ldots n$.  
We compute $\EMD(\bfx^{(1)}, \bfx_0), \ldots \EMD(\bfx^{(n)}, \bfx_0)$ and show them as a box plot.
Doing so for all our perturbation directions $\bfv$ and all natural images $\bfx_0$ produces this figure.
}
\label{fig:emd}
\end{figure}

\begin{figure}[!h]
	\centering
	\begin{subfigure}[t]{0.3\textwidth}
		\centering
		\includegraphics[width=\textwidth]{figures/ssim_boxplot_panda-eps-converted-to.pdf}
		\caption{panda}
	\end{subfigure}
	\begin{subfigure}[t]{0.3\textwidth}
		\centering
		\includegraphics[width=\textwidth]{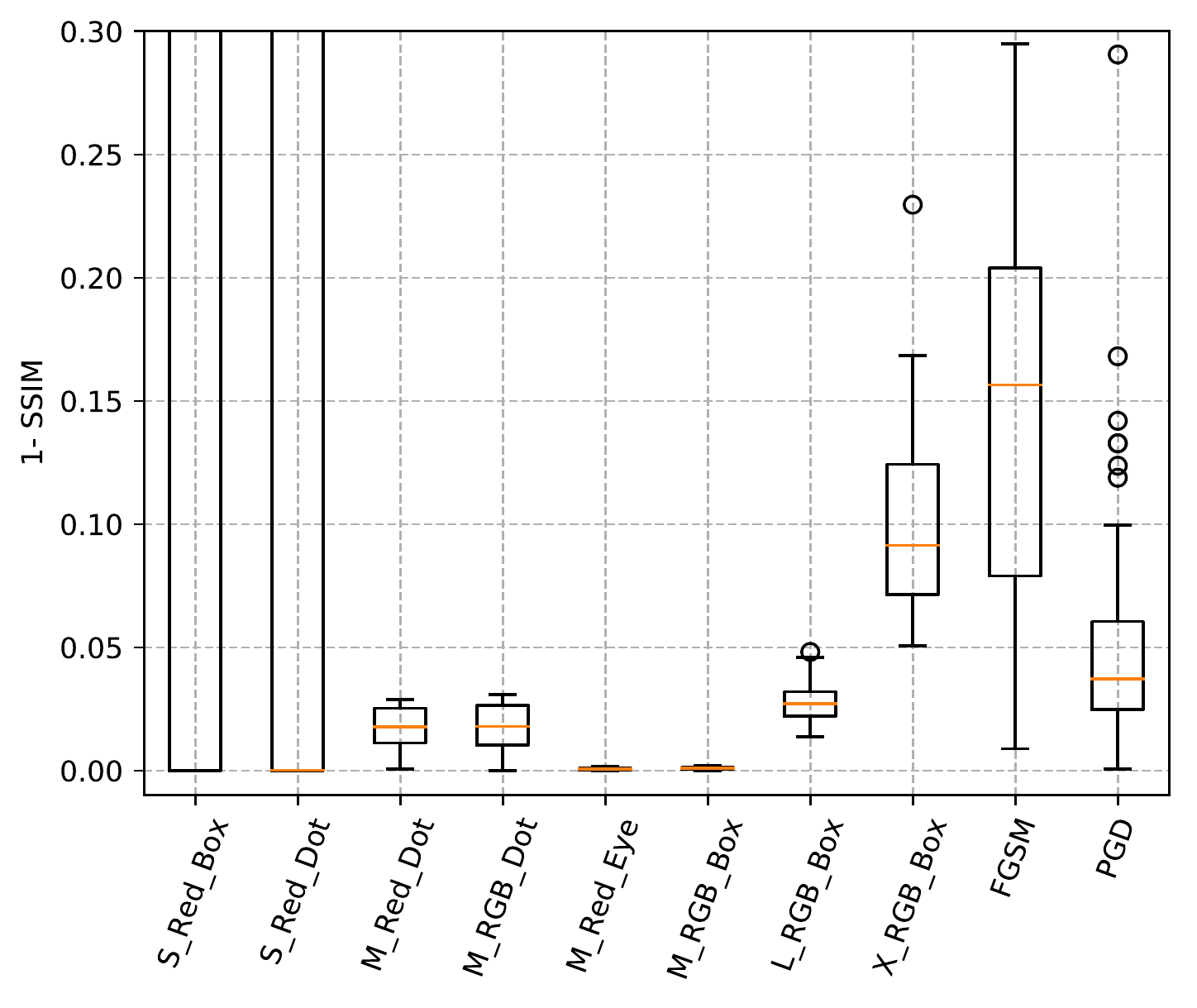}
		\caption{macaw}
	\end{subfigure}
	\begin{subfigure}[t]{0.3\textwidth}
		\centering
		\includegraphics[width=\textwidth]{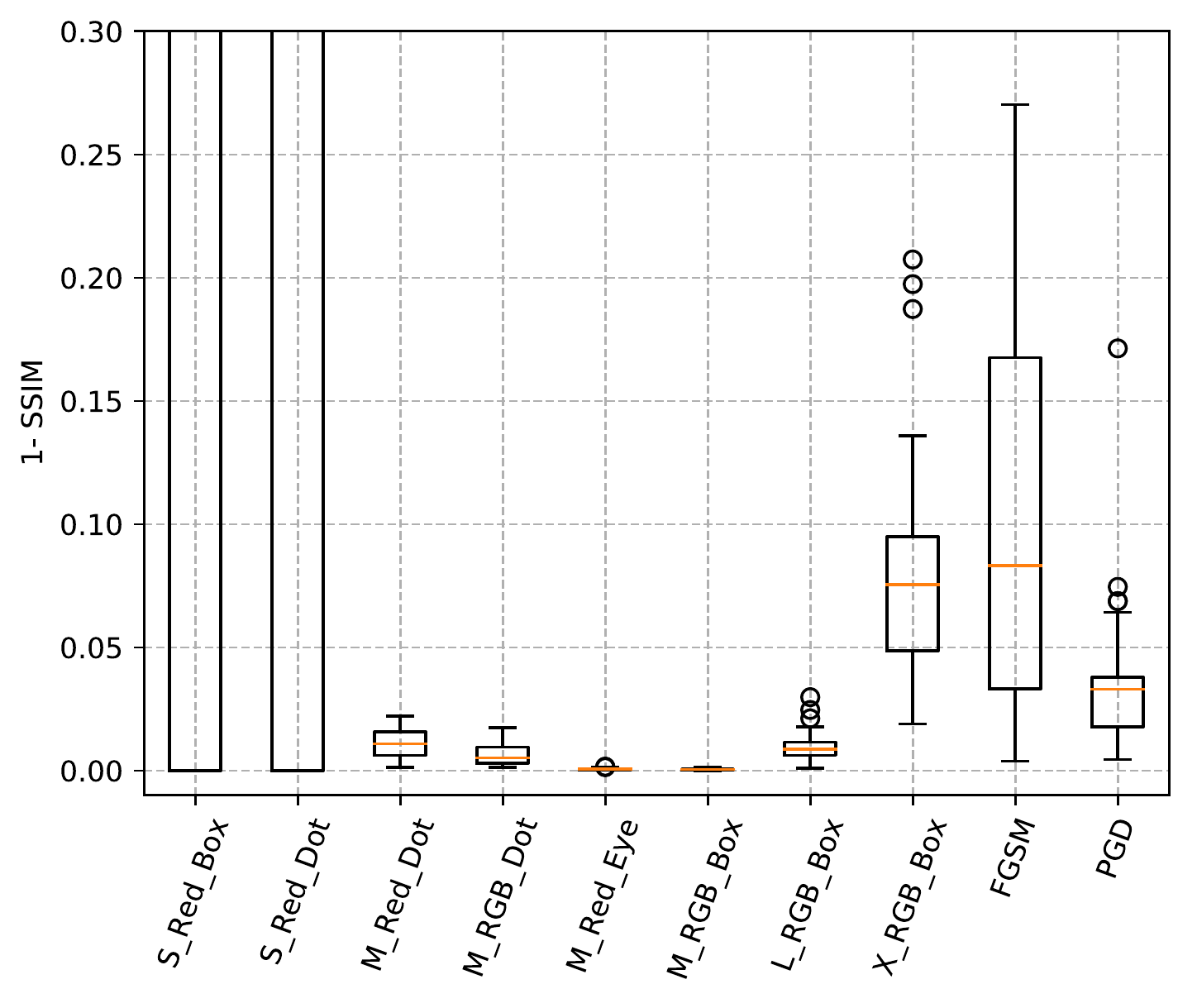}
		\caption{cat}
	\end{subfigure}
	\caption{Box plots of 1 - SSIM on human JND images.}
\label{fig:SSIM}
\end{figure}

\begin{figure}[ht]
\begin{center}
\begin{tabular}{ccc}
panda & macaw & cat \\
		\includegraphics[width=0.25\textwidth]{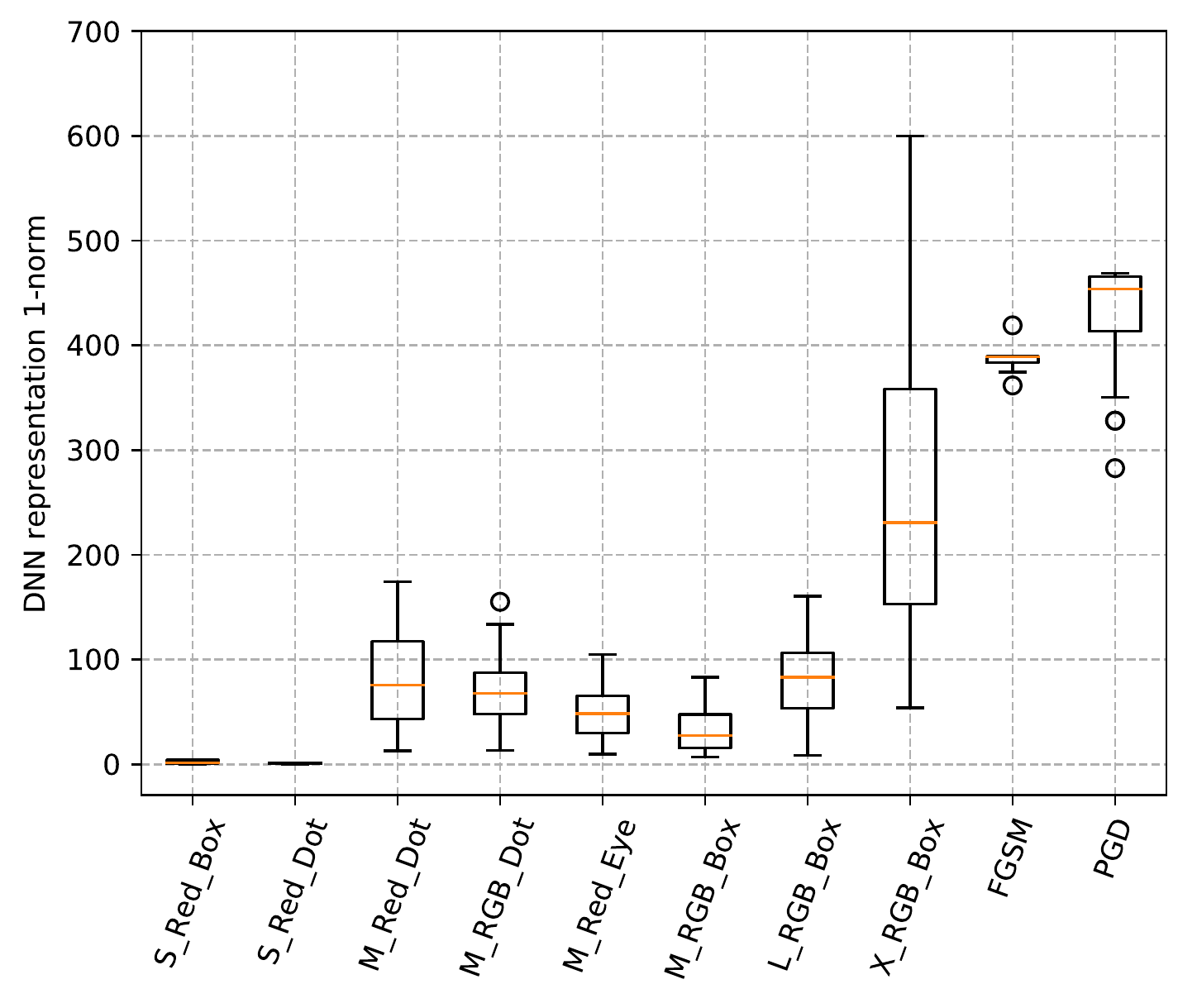} &
		\includegraphics[width=0.25\textwidth]{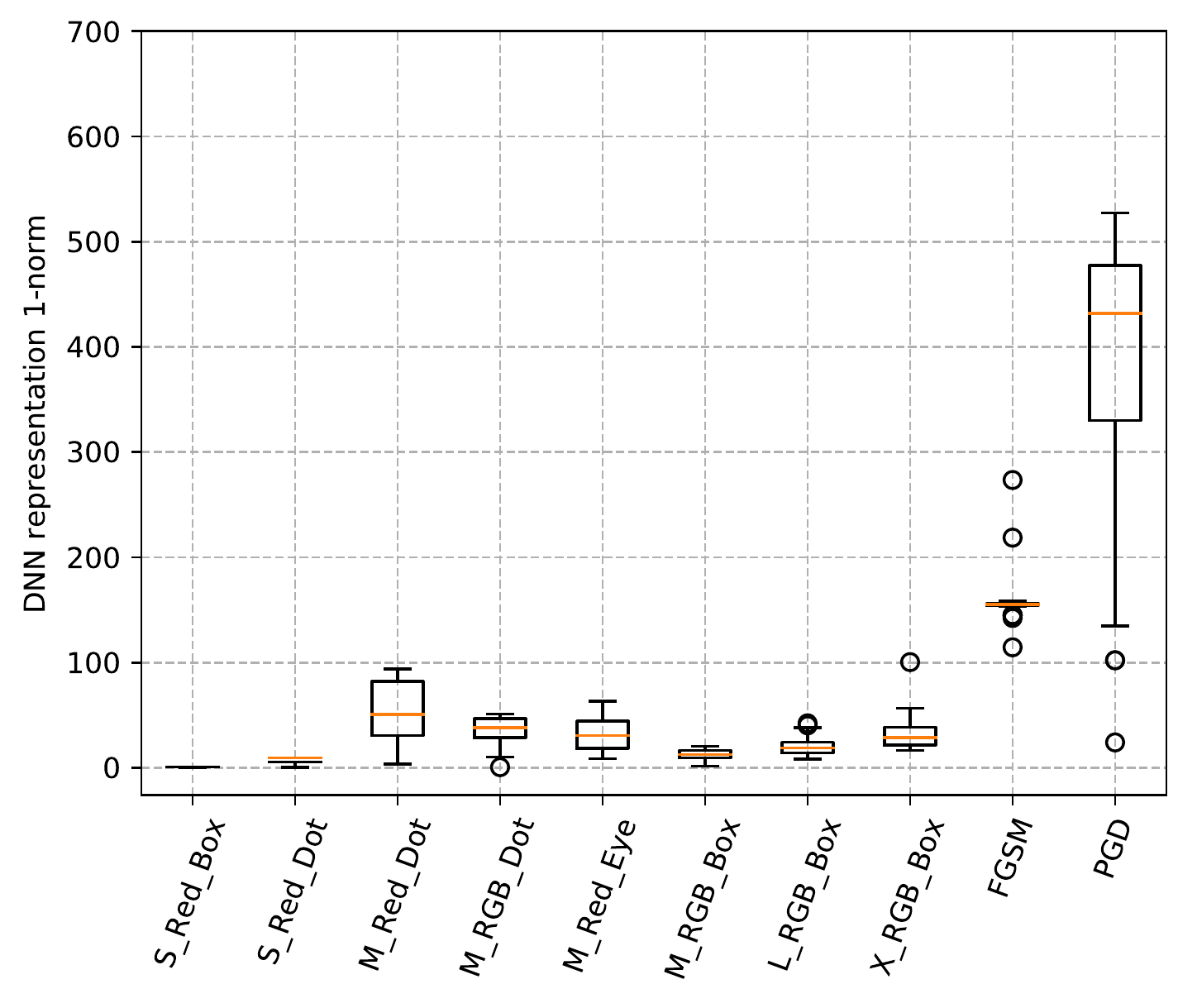} &
		\includegraphics[width=0.25\textwidth]{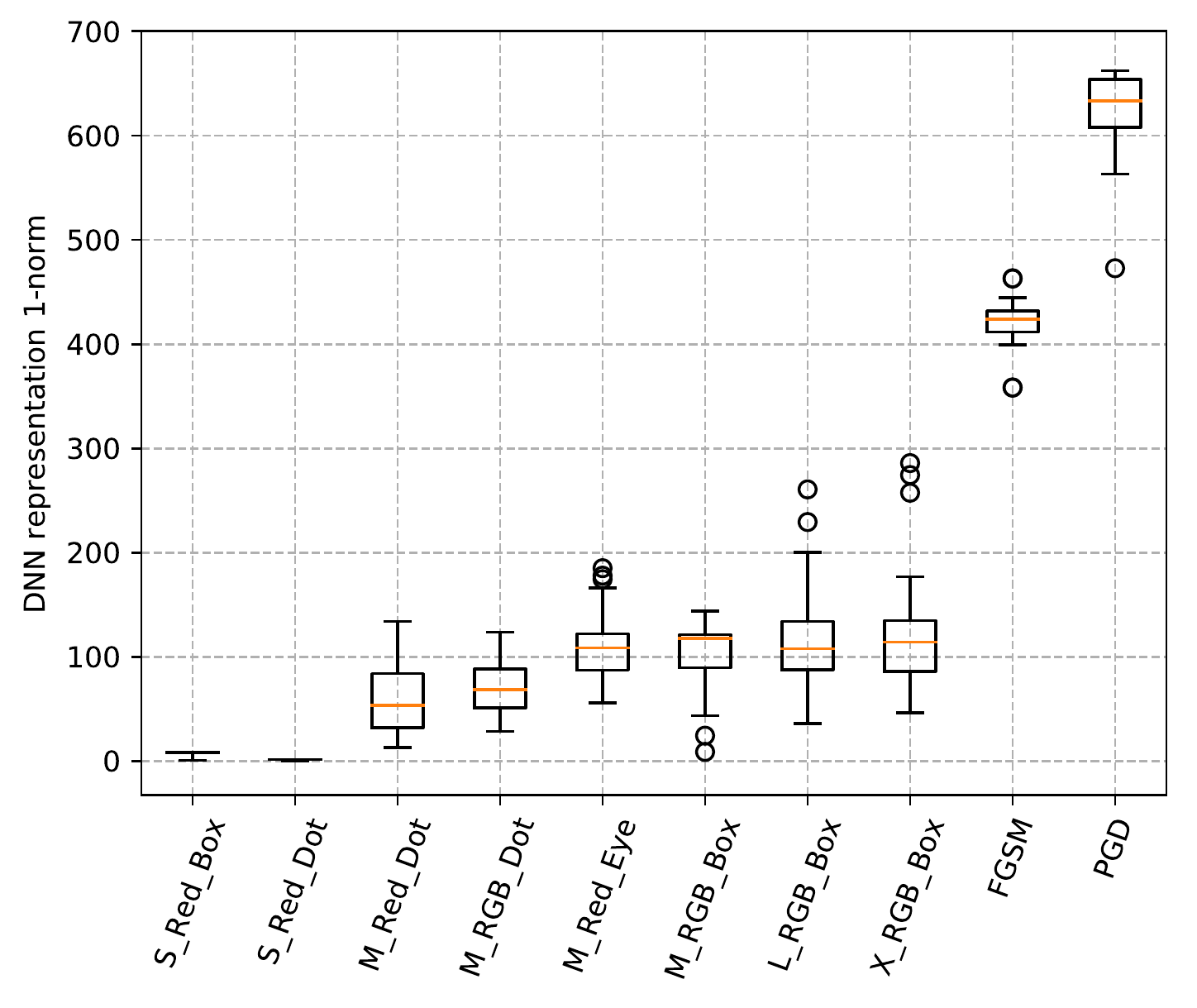}
\\
		\includegraphics[width=0.25\textwidth]{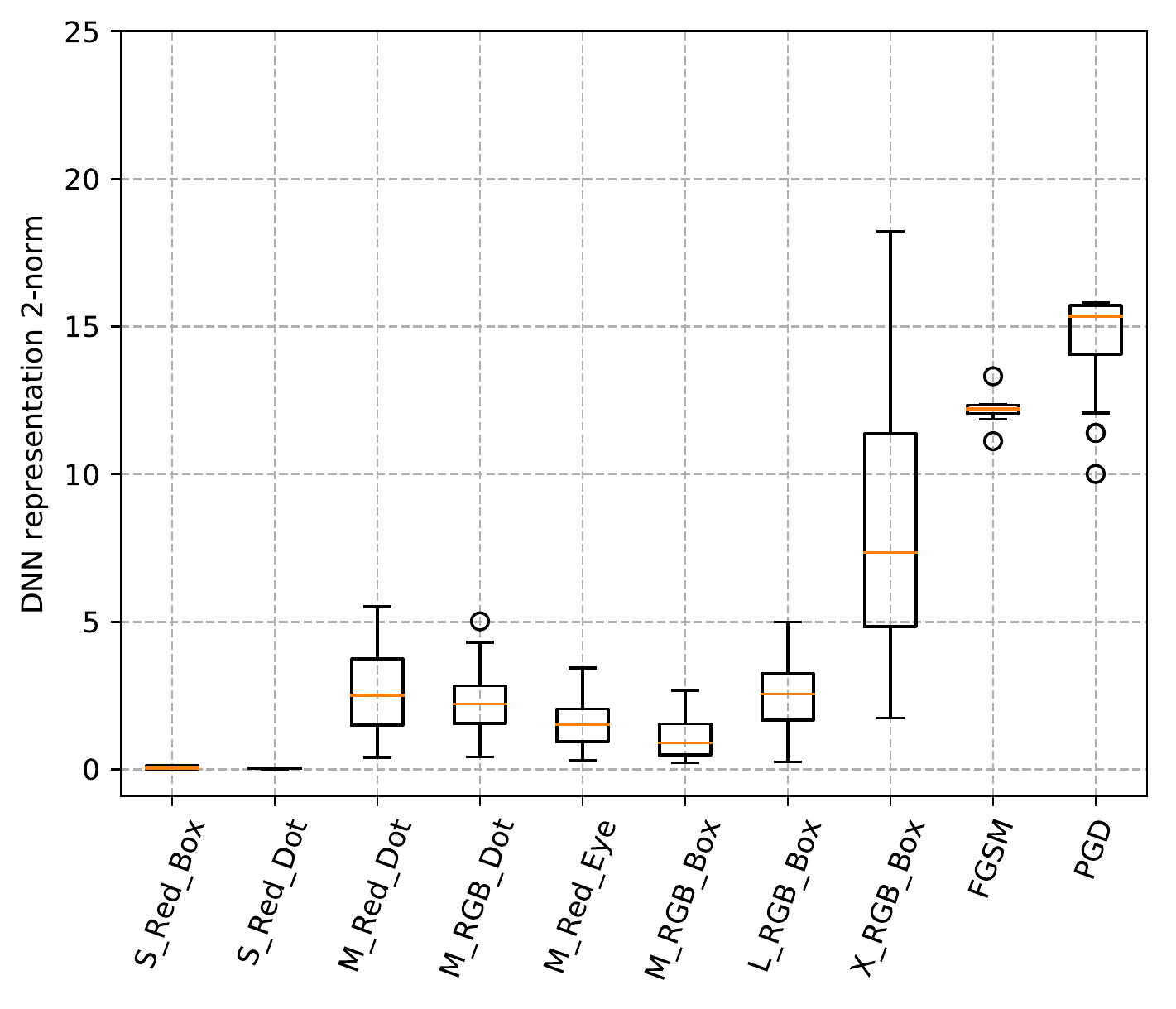} &
		\includegraphics[width=0.25\textwidth]{figures/embedding_2_norm_boxplot_macaw-eps-converted-to.pdf} &
		\includegraphics[width=0.25\textwidth]{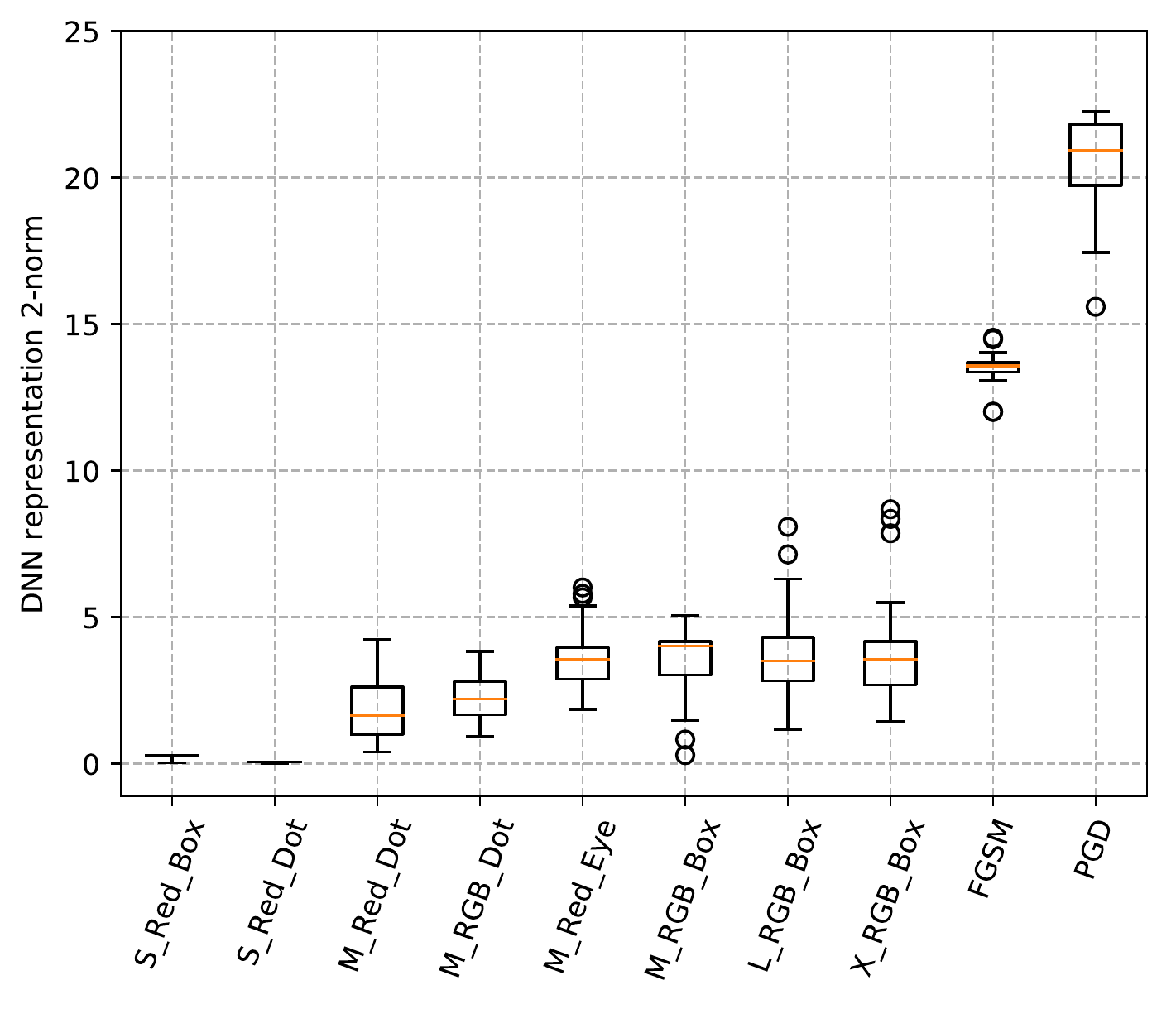}
\\
		\includegraphics[width=0.25\textwidth]{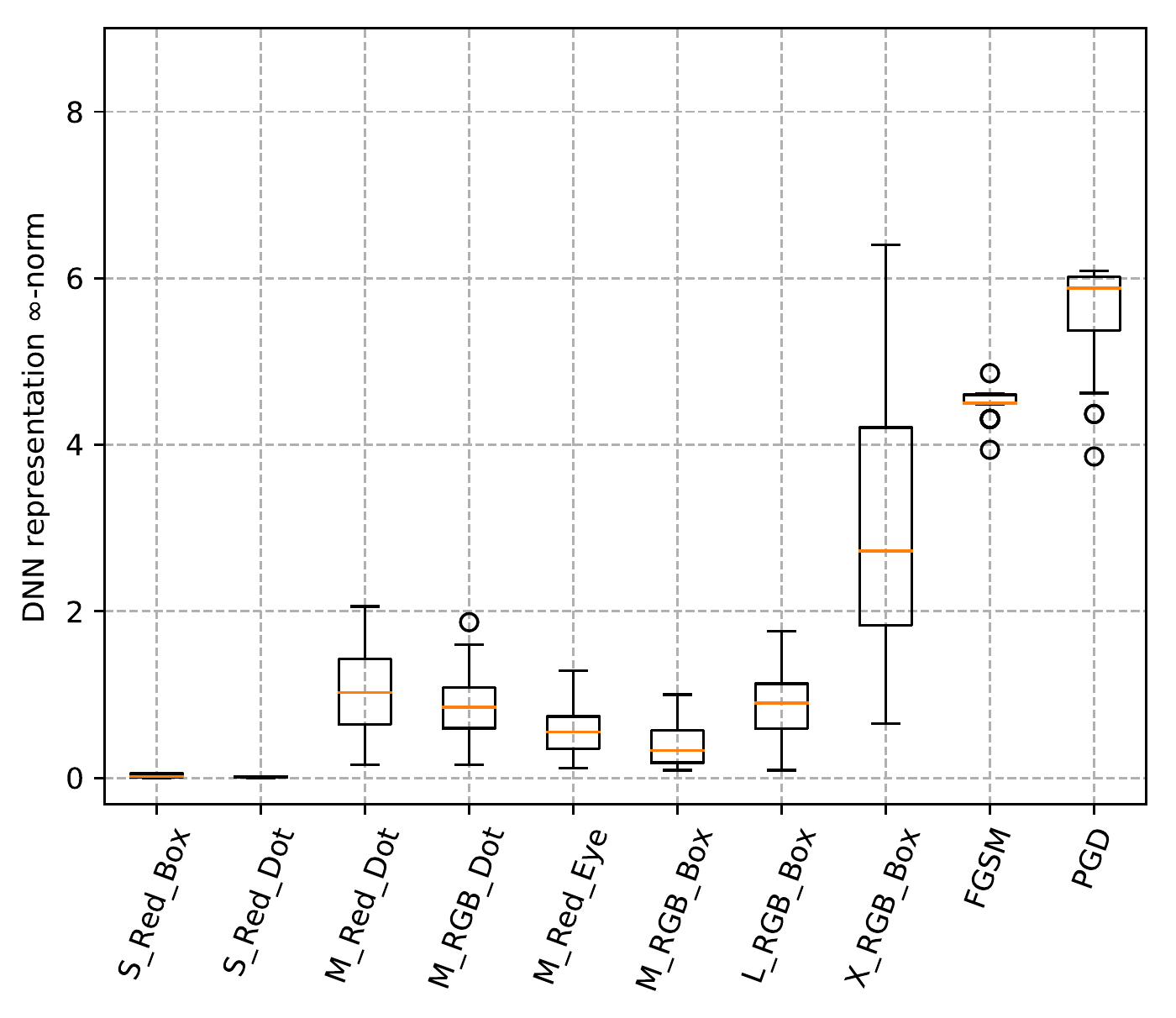} &
		\includegraphics[width=0.25\textwidth]{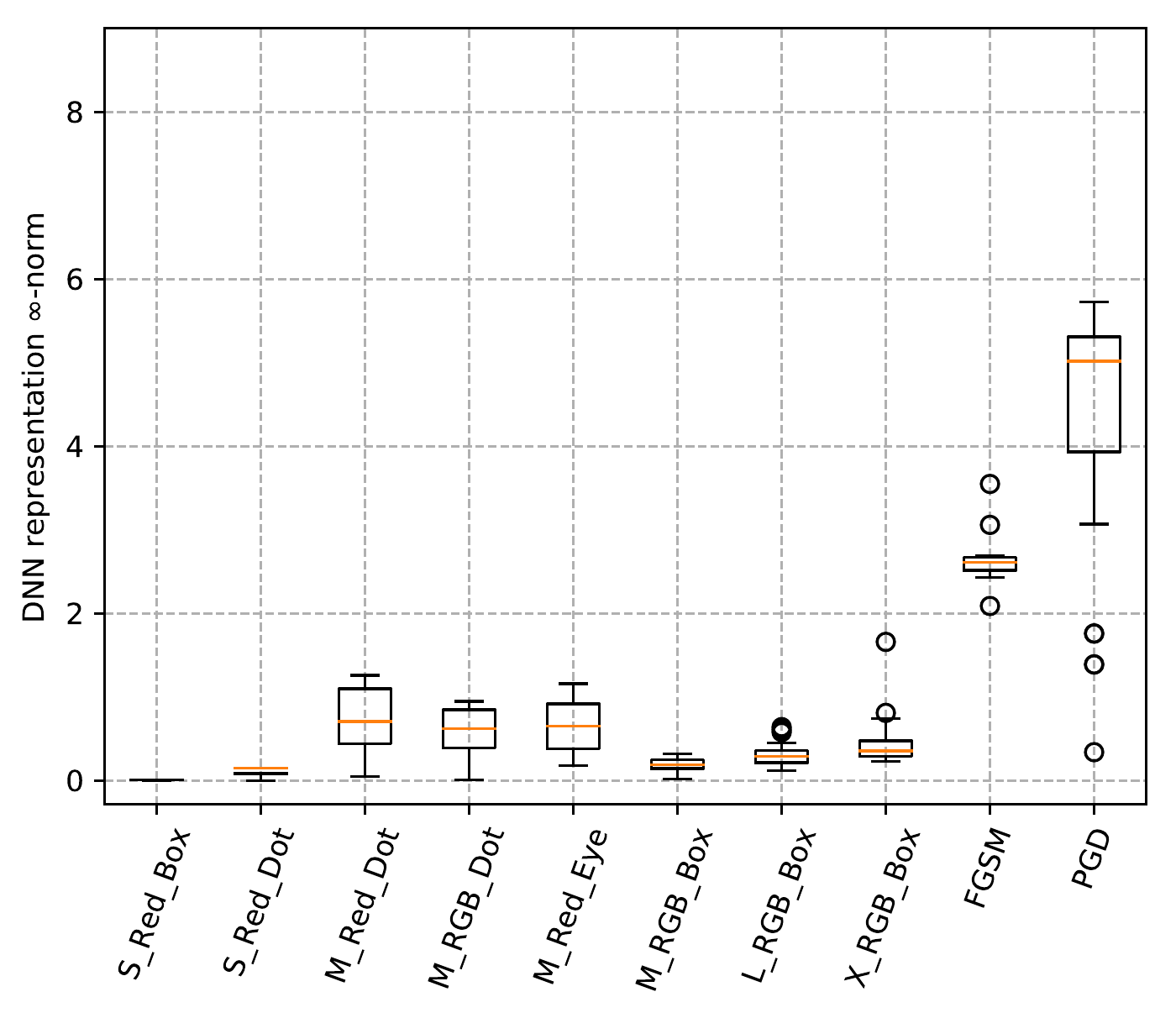} &
		\includegraphics[width=0.25\textwidth]{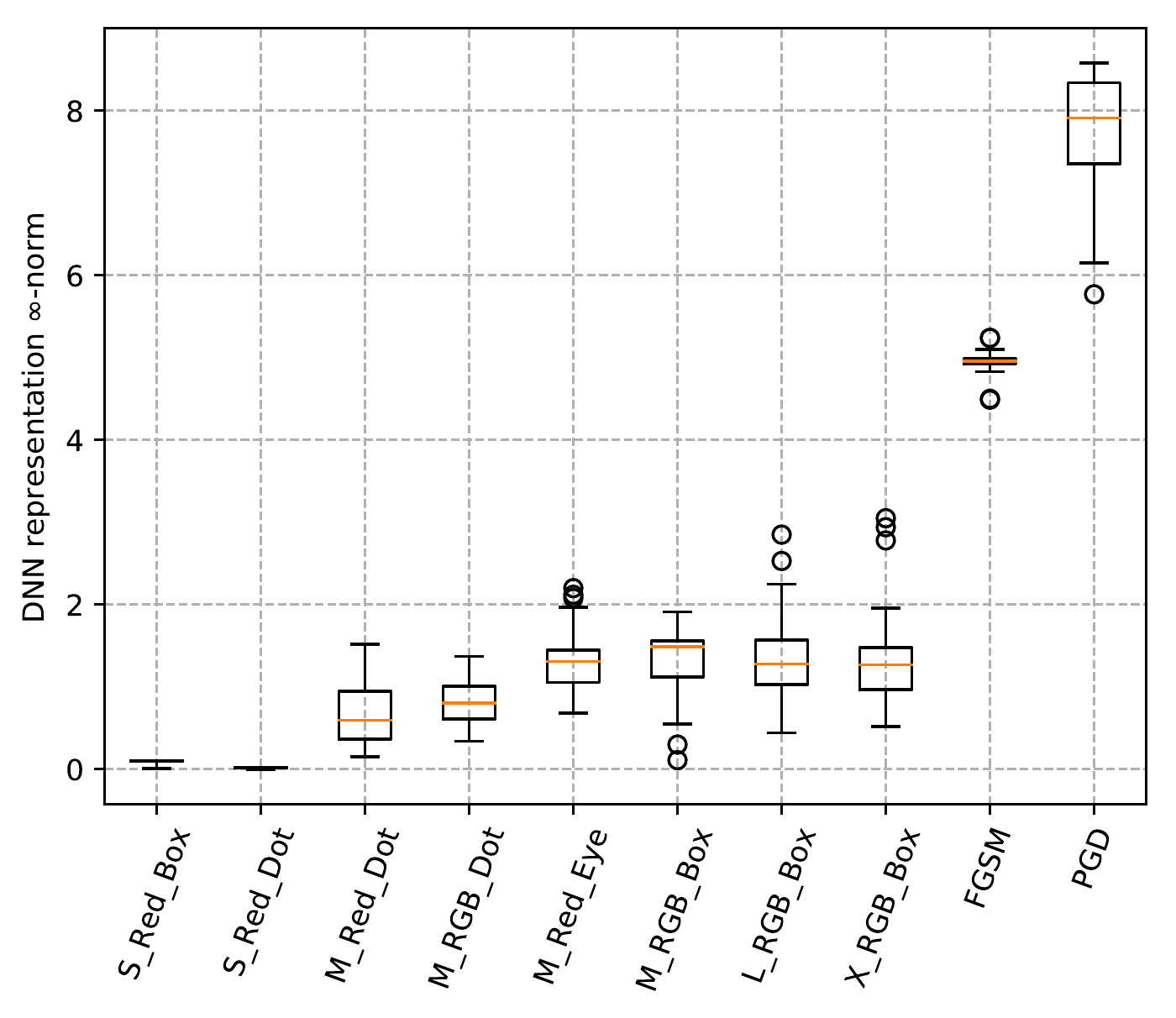}
\end{tabular}
\end{center}
	\caption{Box plots of DNN $\|\xi(\bfx) - \xi(\bfx_0)\|_p$ on human JND images.  rows: $p=1,2,\infty$, respectively.
}
\label{fig:embedding_Lp}
\end{figure}
\begin{figure}[htb]
\centering
\includegraphics[width=0.8\textwidth]{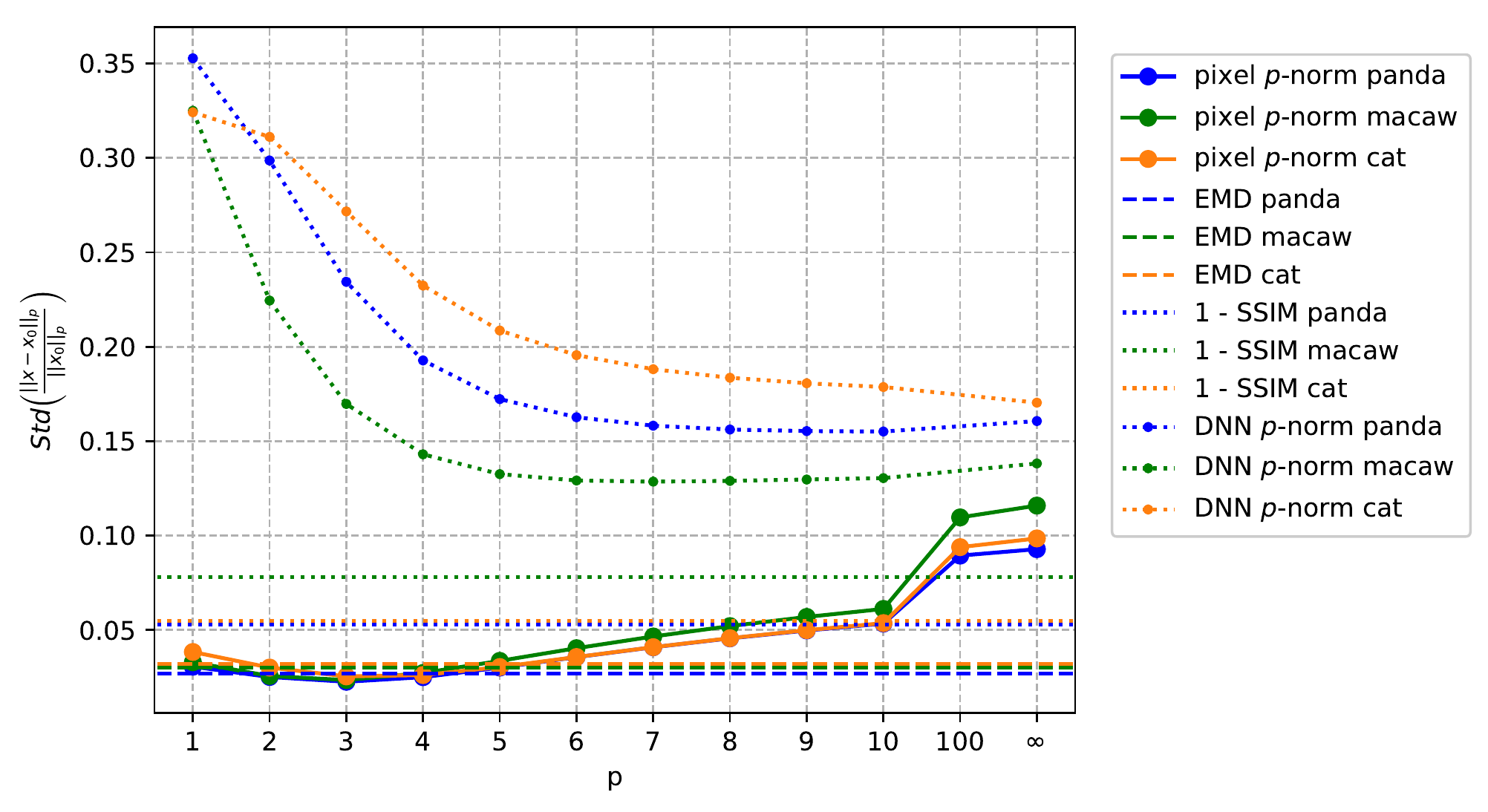}
\caption{Variability of fit to human data including DNN representations, lower is better}
\label{fig:varDNN}
\end{figure}

\FloatBarrier

\subsection{Amazon Mechanical Turk instructions}
For reference, screenshots of the instructions displayed to participants are included in this appendix.
\begin{figure}[H]
  \centering
  \includegraphics[width=.9\textwidth]{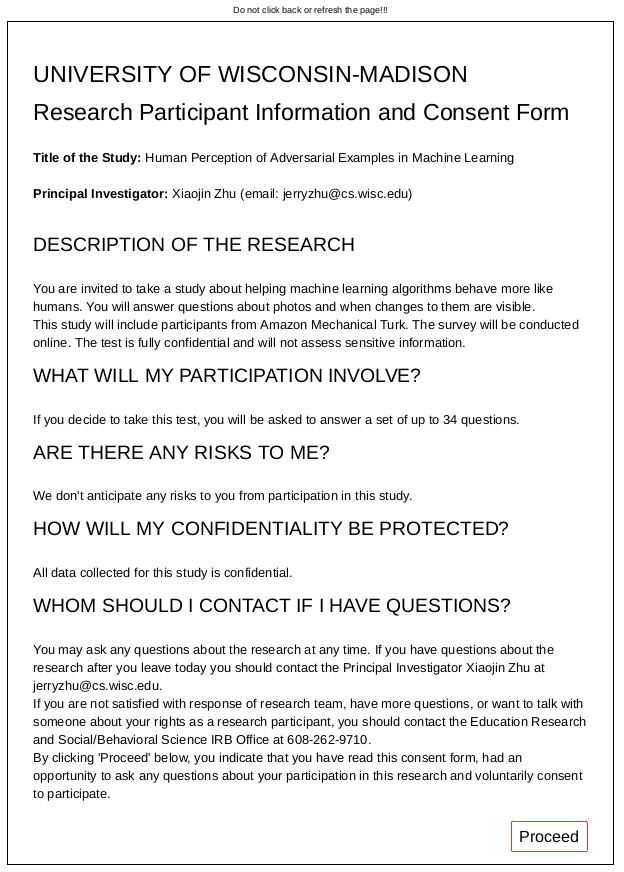}
  \caption{Instruction Page 1}
\end{figure}

\begin{figure}[H]
  \centering
  \begin{minipage}{.5\textwidth}
    \centering
    \includegraphics[width=\linewidth]{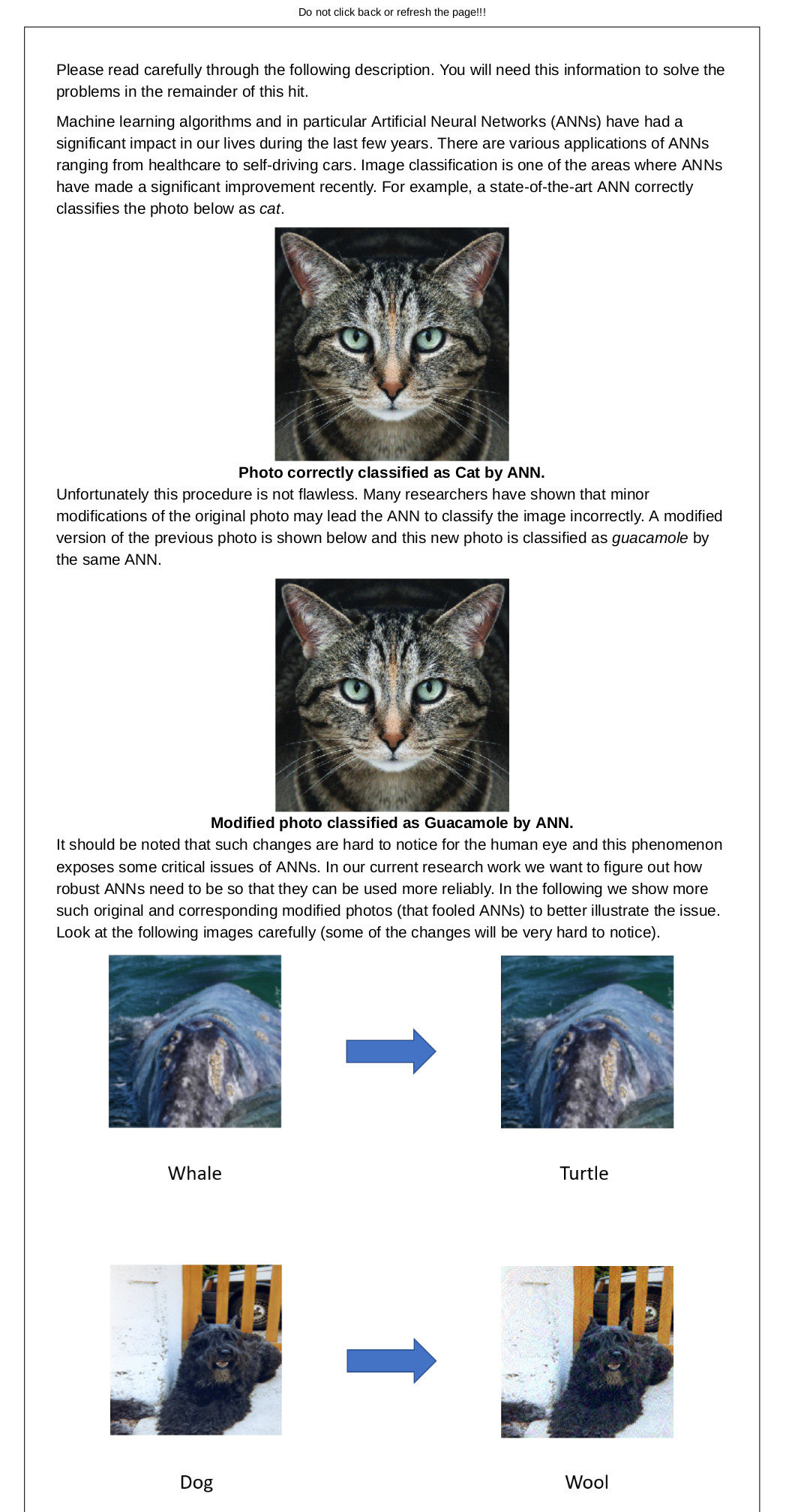}
  \end{minipage}%
  \begin{minipage}{.5\textwidth}
    \centering
    \includegraphics[width=\linewidth]{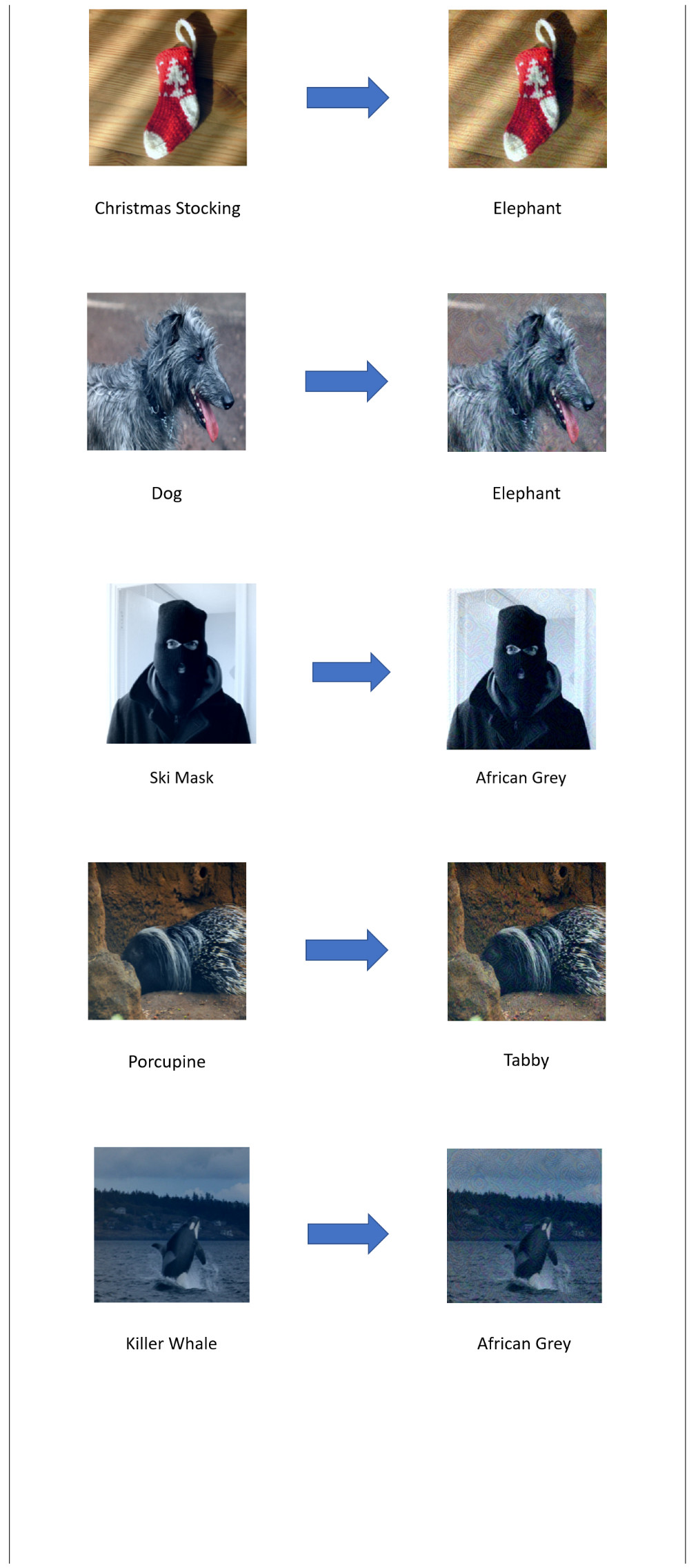}
  \end{minipage}
  \caption{Instruction Page 2}
\end{figure}

\begin{figure}[H]
  \centering
  \begin{minipage}{.5\textwidth}
    \centering
    \includegraphics[width=\linewidth]{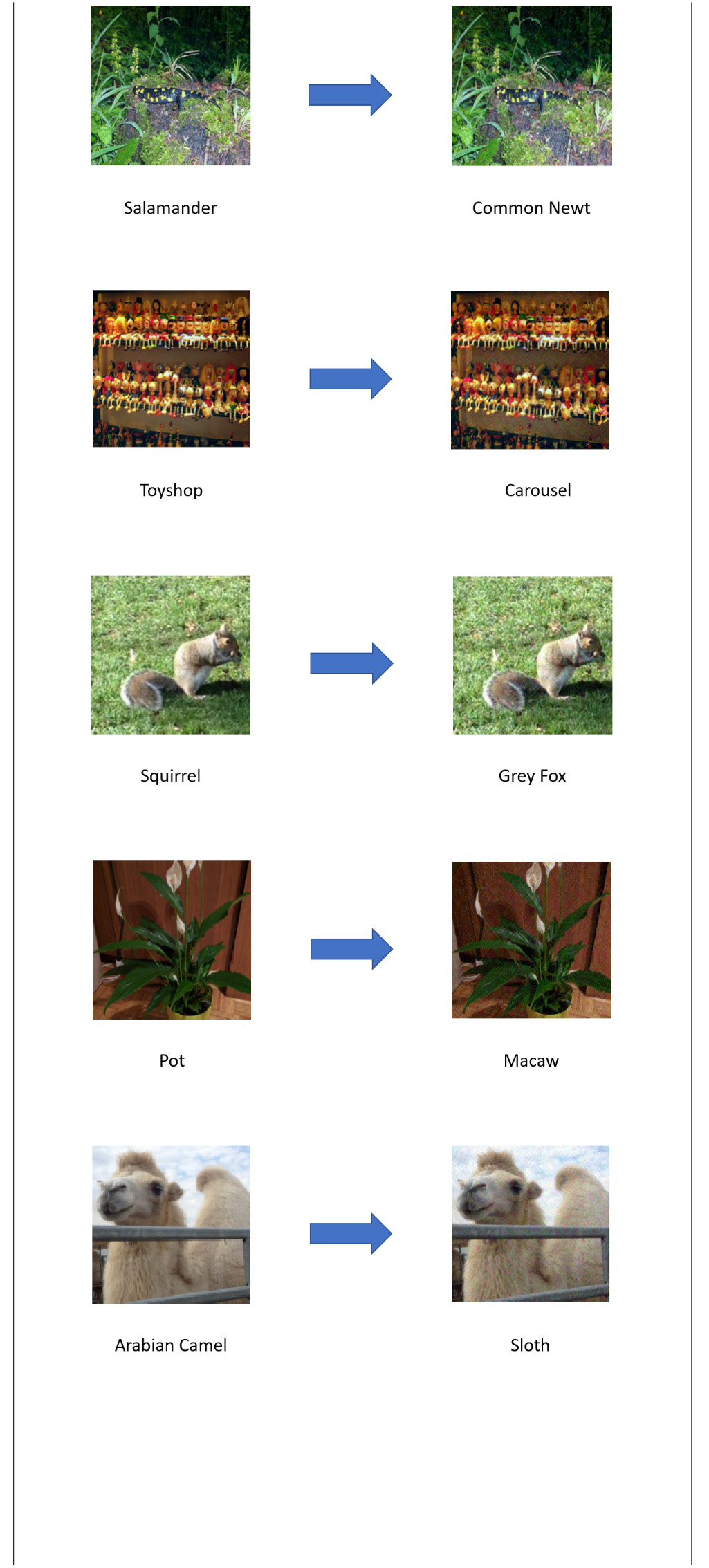}
  \end{minipage}%
  \begin{minipage}{.5\textwidth}
    \centering
    \includegraphics[width=\linewidth]{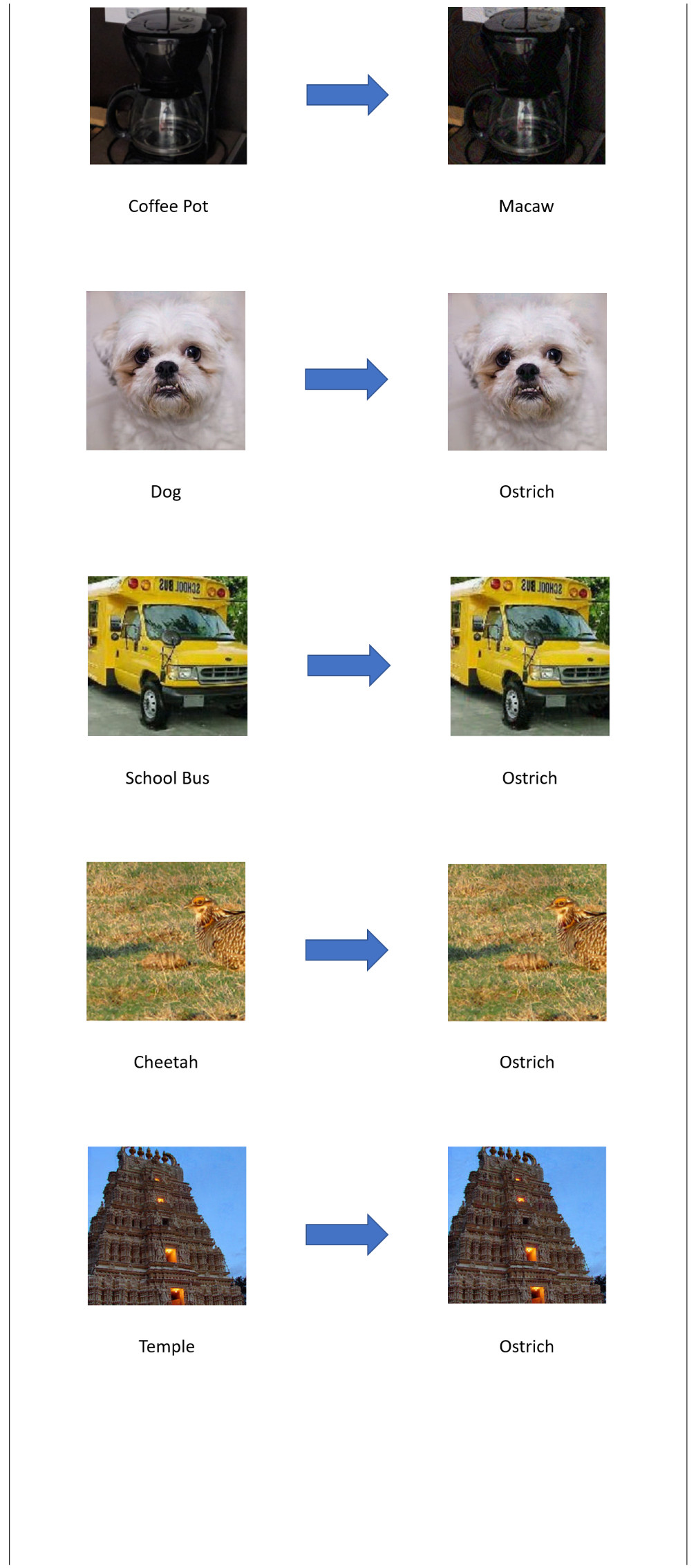}
  \end{minipage}
  \caption{Instruction Page 2 (\textit{cont'd})}
\end{figure}
\begin{figure}[H]
  \centering
  \includegraphics[width=\linewidth]{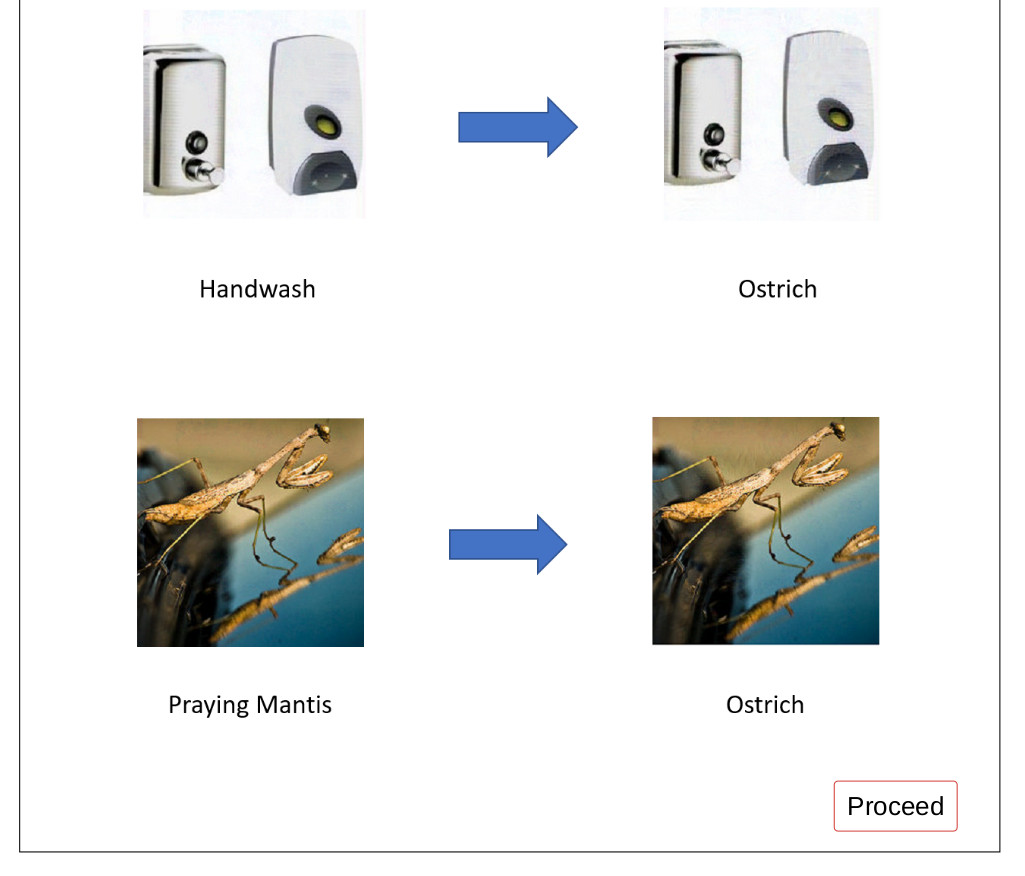}
  \caption{Instruction Page 2 (\textit{cont'd})}
\end{figure}

\begin{figure}[H]
  \centering
  \includegraphics[width=.6\textwidth]{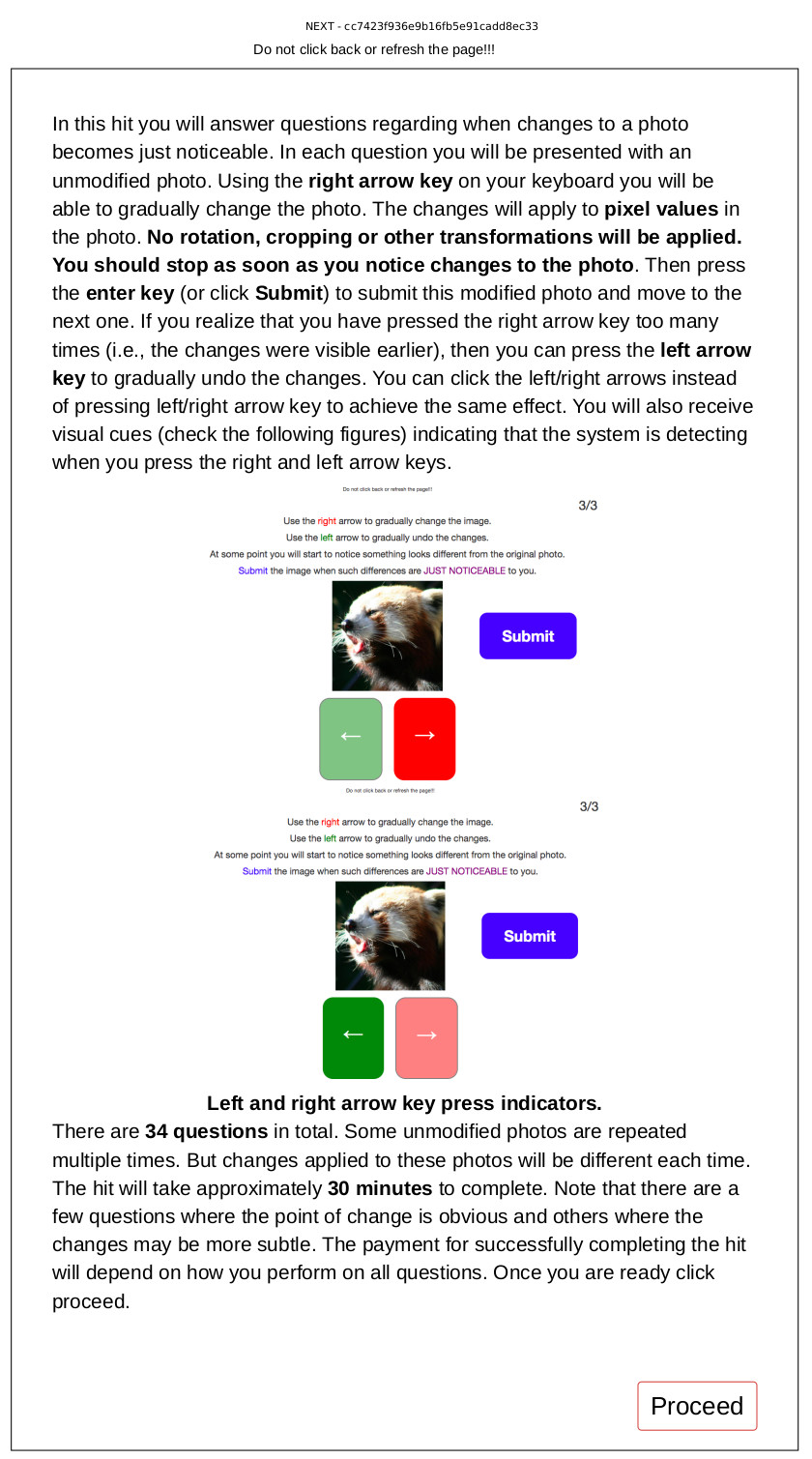}
  \caption{Instruction Page 3}
\end{figure}

\end{document}